\pdfoutput=1

\documentclass[11pt]{article}

\usepackage[final]{acl}

\usepackage{times}
\usepackage{latexsym}

\usepackage[T1]{fontenc}

\usepackage[utf8]{inputenc}

\usepackage{microtype}

\usepackage{inconsolata}

\usepackage{graphicx}
\usepackage{booktabs}
\usepackage{multirow}
\usepackage[table]{xcolor}
\usepackage{adjustbox}
\usepackage{amsthm}
\newtheorem{theorem}{Theorem}

\usepackage[ruled,vlined]{algorithm2e}
\usepackage[font=small]{caption}
\usepackage{subcaption}
\usepackage{amsfonts}
\usepackage{amsmath}
\usepackage{array}
\usepackage{tikz}
\usepackage{newfloat}
\usepackage{listings}
\usepackage{mathtools}
\usepackage{amssymb}

\theoremstyle{definition}
\newtheorem{definition}{Definition}
\theoremstyle{remark}

\newtheorem{proposition}{Proposition}
\newtheorem{corollary}{Corollary}
\usepackage{enumitem}
\usepackage{pifont}
\newcommand{\cmark}{{\color[HTML]{2E7D32}\ding{51}}}
\newcommand{\xmark}{{\color[HTML]{C62828}\ding{55}}}
\newcommand{\sd}[1]{{\tiny\textcolor{gray}{±#1}}}
\usepackage{threeparttable}
\definecolor{blue}{HTML}{3331D7}
\definecolor{orange}{HTML}{E57932}
\definecolor{purple}{HTML}{663399}
\definecolor{green}{HTML}{41805E}
\definecolor{rose}{HTML}{C71585}
\definecolor{crimson}{HTML}{DC143C}
\definecolor{grey}{HTML}{505050}
\definecolor{skyblue}{RGB}{203, 221, 245}
\definecolor{lightorange}{RGB}{255, 243, 224}
\definecolor{lightpurple}{RGB}{232, 222, 245}

\title{Gradients Know What Outcomes Don't: Unlocking Reinforcement Learning for LLM Reasoning with Gradient-Aligned Rewards}

\author{
 \textbf{Leqi Zheng \textsuperscript{1,\textdagger}},
 \textbf{Jinbo Su\textsuperscript{2,\textdagger}},
 \textbf{Fang Niu \textsuperscript{1,\textdagger}},
 \textbf{Chaokun Wang* \textsuperscript{1}},
 \\
 \textbf{Weiping Wang \textsuperscript{3}},
\textbf{Jiajun Zhang \textsuperscript{4}},
\textbf{Shannan Yan \textsuperscript{1}},
\\
\textbf{Jie Wu \textsuperscript{5}},
\textbf{Zhaolu Kang \textsuperscript{6}},
\textbf{Rong Fu \textsuperscript{7}},
\textbf{Hang Zhang \textsuperscript{1}}
\\
 \textsuperscript{1}Tsinghua University, 
 \textsuperscript{2}Renmin University of China, 
 \textsuperscript{3}Institute of Information Engineering, CAS,
 \\
 \textsuperscript{4}USTC,
\textsuperscript{5}The Australian National University,
 \textsuperscript{6}Peking University,
 \textsuperscript{7}University of Macau
\\
 \small{\textsuperscript{\textdagger} Equal contribution.}
 \\
 \small{\textbf{* Correspondence:} {chaokun@tsinghua.edu.cn}}
}

\begin{document}
\maketitle

\begin{abstract}
Reinforcement learning from verifiable rewards (RLVR) drives chain-of-thought reasoning in large language models, yet its binary outcome reward cannot distinguish among correct trajectories.
Existing dense reward alternatives, from surface heuristics to process reward models, either ignore the expert solutions already present in training corpora or require expensive offline annotation.
We propose \textbf{Gradient-Aligned Reward (GAR)}, which operates in the policy's own gradient space: truncated backpropagation through the output projection layer extracts a compact gradient vector for each rollout, and cosine similarity with an expert-anchor gradient yields a dense, reasoning-aware reward with less than 9\% wall-clock overhead.
We prove that this cosine admits a multiplicative decomposition into prediction-error and activation-pattern factors, providing a concrete characterization of what the alignment signal measures.
On Qwen3-4B and Qwen3-8B, GAR consistently improves over GRPO and other baselines on competition-level math benchmarks and transfers to GPQA Diamond and MMLU-Pro without domain-specific data.
Code and data are available at \url{https://github.com/LQgdwind/GAR}.

\end{abstract}
\section{Introduction}
\label{sec:intro}
\begin{figure}[t]
\centering
\includegraphics[width=\columnwidth]{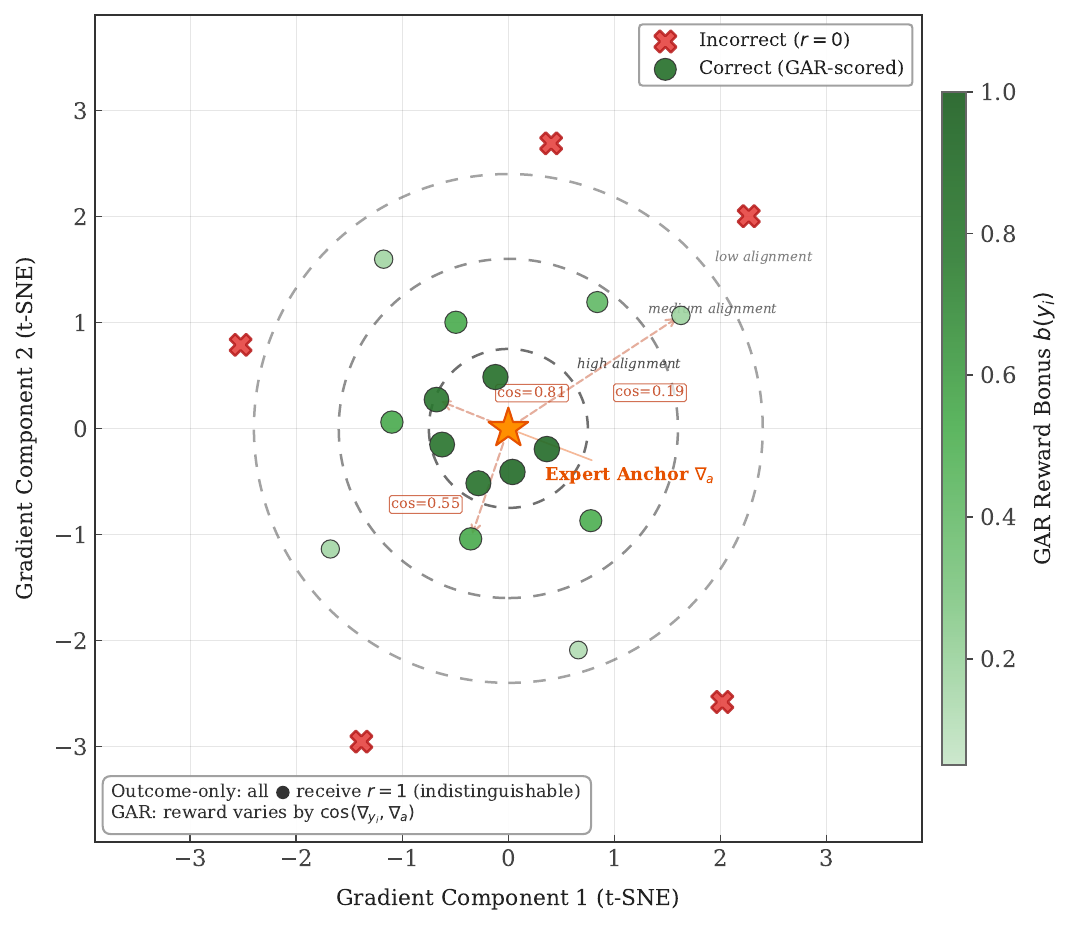}
\caption{Gradient-space visualization of GAR. The expert anchor $\mathbf{v}_a$ (star) defines the reference direction in the parameter update manifold. Correct rollouts (circles) are scored by cosine similarity with the anchor: high-alignment rollouts (dark) receive a large GAR bonus, while low-alignment rollouts (light) receive a smaller bonus despite also being correct. Incorrect rollouts (crosses) are gated out and receive zero reward regardless of their gradient direction.}
\label{fig:gradient_space}
\end{figure}

Reinforcement learning from verifiable rewards (RLVR) has emerged as a compelling paradigm for eliciting chain-of-thought reasoning in large language models without supervised fine-tuning~\cite{guo2025deepseek,shao2024deepseekmath,yu2025dapo,NEURIPS2025_9a92ea37,zheng2026should,zheng-etal-2025-lagcl4rec}.
Under this setting, a base model is optimized solely with a binary outcome reward that verifies final-answer correctness, and DeepSeek-R1-Zero~\cite{guo2025deepseek} demonstrated that structured reasoning~\cite{wei2022chain,kojima2022large} can emerge spontaneously from such sparse supervision.
Despite this success, the exclusive reliance on binary rewards introduces a fundamental \emph{credit assignment} problem: once multiple rollouts produce correct answers, the reward signal becomes flat over the correct subset, and the resulting policy gradient carries no information to preferentially reinforce higher-quality reasoning trajectories.

This pathology motivates a search for denser, process-level supervision, yet existing remedies each suffer from significant limitations:
(1)~\textbf{Expert solutions remain unused.}
Outcome-only RLVR assigns identical reward to every correct response, collapsing the group-relative advantage among the correct subset to zero~\cite{shao2024deepseekmath}.
Widely used math corpora such as NuminaMath-CoT~\cite{li2024numinamath} already ship expert chain-of-thought solutions alongside each problem, yet no existing reward mechanism exploits them to provide a process-level training signal.
(2)~\textbf{Surface-level heuristic shaping.}
Rule-based reward shaping methods~\cite{wen2025light,aggarwal2025l1} introduce length or format penalties as proxies for reasoning quality, but such heuristics operate on surface attributes and cannot evaluate whether the underlying chain of thought genuinely engages the reasoning structures required by the task.
(3)~\textbf{Expensive and offline process supervision.}
Process reward models (PRMs) provide step-level feedback but require large-scale expert annotations~\cite{lightman2023let,uesato2022solving}; even automated alternatives~\cite{wang2024math,cui2025process,setlur2024rewarding} are trained offline on fixed distributions that diverge from the evolving policy.
A detailed comparison is provided in Table~\ref{tab:comparison}.
\begin{table}[t]
\centering
\small
\caption{Comparison of reward paradigms. GAR is the only method that leverages expert CoT in gradient space.}
\resizebox{\columnwidth}{!}{%
\begin{tabular}{lcccc}
\toprule
\rowcolor{lightpurple} \textbf{Method} & \textbf{Dense Sig.} & \textbf{Expert CoT} & \textbf{Grad.-Space} & \textbf{No Extra Model} \\
\midrule
GRPO          & \xmark & \xmark & \xmark & \cmark \\
Grad2Reward   & \cmark & \xmark & \cmark & \xmark \\
G2RL          & \cmark & \xmark & \cmark & \cmark \\
\rowcolor{lightorange} \textbf{GAR (Ours)} & \cmark & \cmark & \cmark & \cmark \\
\bottomrule
\end{tabular}%
}
\label{tab:comparison}
\end{table}
These limitations point to a natural question: \textit{can we turn the expert solutions that training corpora already provide into a gradient-space signal that differentiates among correct rollouts?}

We find that we can, and introduce \textbf{Gradient-Aligned Reward (GAR)}, a lightweight online process reward mechanism for RLVR.
The central insight is that if two correct trajectories implement substantively similar reasoning, their gradient directions through the output projection layer must remain close, irrespective of surface-level textual variation.
GAR operationalizes this hypothesis by performing truncated backpropagation through only the LM head and scoring each rollout by its cosine similarity with an expert-anchor gradient derived from existing chain-of-thought solutions in the training corpus (Figure~\ref{fig:gradient_space}).
This design addresses all three limitations: gradient cosine with expert anchors turns readily available CoT solutions into a dense reward; the gradient signal captures reasoning-level structure beyond surface attributes; and the online, truncated computation eliminates dependence on step-level annotations or external reward models.

In summary, our contributions are four-fold:

(1)~\textbf{Gradient-Space Process Rewards from Expert Anchors.}
We introduce a reward mechanism that converts readily available expert chain-of-thought solutions into gradient-space reference vectors, enabling dense, per-rollout process supervision without additional annotation.
We ground the signal via the empirical neural tangent kernel and potential-based reward shaping theory.

(2)~\textbf{Multiplicative Decomposition of the Alignment Signal.}
We prove that the gradient cosine decomposes multiplicatively into a prediction-error factor and an activation-pattern factor (Theorem~\ref{thm:decomposition}), providing a concrete characterization of the two complementary axes along which GAR distinguishes correct trajectories.

(3)~\textbf{Lightweight Online Computation.}
Truncated backpropagation through the output projection layer reduces alignment cost to $O(V \times d)$, adding less than 9\% wall-clock overhead to standard GRPO training.

(4)~\textbf{Empirical Validation.}
On Qwen3-4B and 8B base models, GAR yields consistent pass@$k$ gains on four competition-level math benchmarks and transfers to GPQA Diamond and MMLU-Pro without domain-specific training data.

\section{Method}
\label{sec:method}
\label{sec:flat-reward-pathology}

\begin{figure*}[t]
\centering
\vspace{-2mm}
\includegraphics[width=\textwidth]{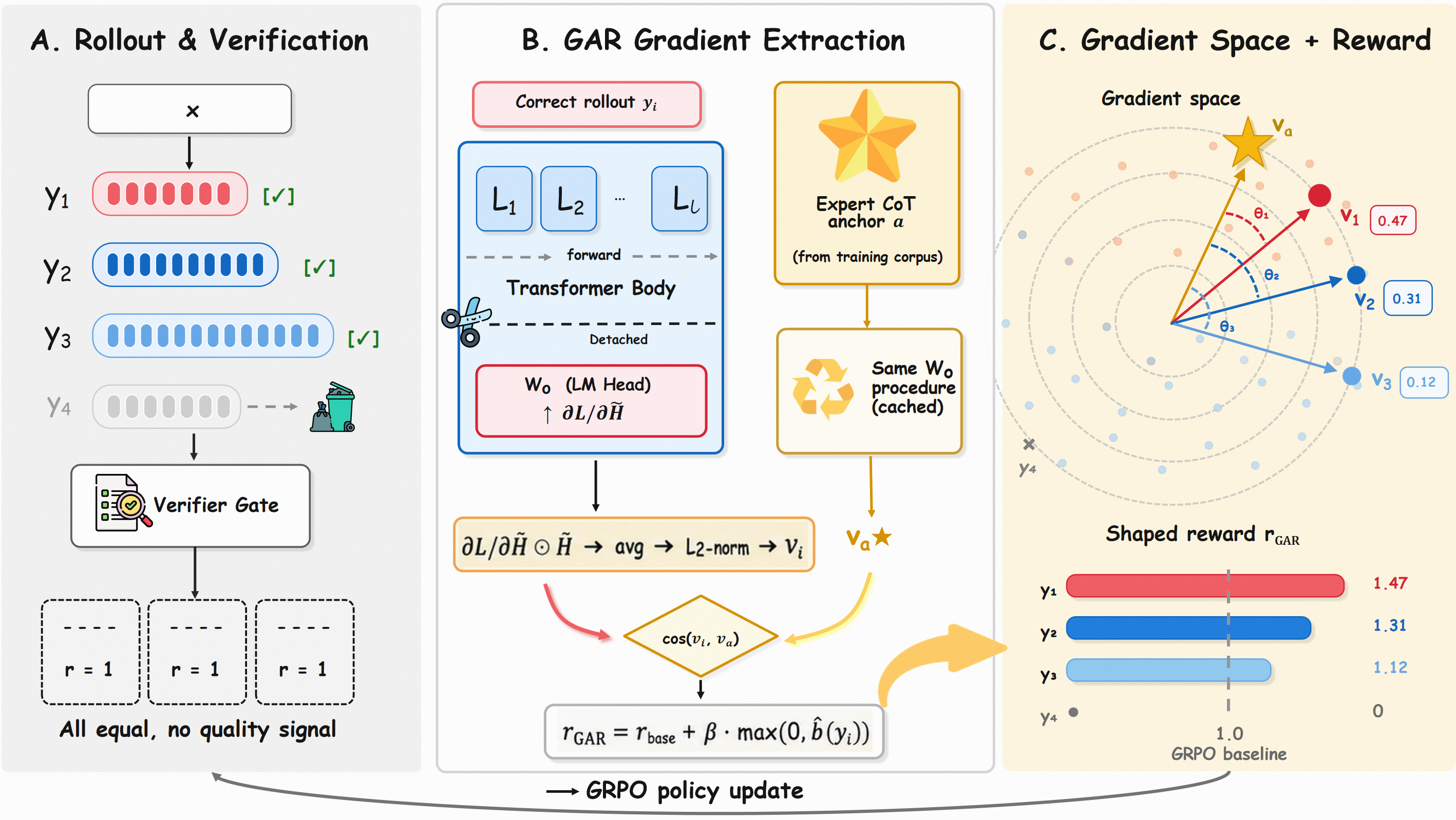}
\vspace{-2mm}
\caption{GAR pipeline within one GRPO step. \textbf{(A)}~The policy generates $K$ rollouts; a verifier gates correct responses (all receiving $r{=}1$ under standard GRPO). \textbf{(B)}~For each correct rollout, a teacher-forcing forward pass produces hidden states $\widetilde{\mathbf{H}}$, which are detached at the output projection boundary and backpropagated through $W_o$ only to obtain gradient-activation vector $\mathbf{v}_i$. The expert CoT undergoes the same procedure (cached) to yield $\mathbf{v}_a$; their cosine similarity gives the alignment bonus. \textbf{(C)}~Rollouts whose gradient directions cluster near $\mathbf{v}_a$ receive higher rewards, breaking the flat-reward pathology.}
\vspace{-2mm}
\label{fig:backbone}
\end{figure*}

GAR is motivated by a fundamental limitation of outcome-only reinforcement learning.
Under a binary verifier reward $r(x, y) = \mathbf{1}[\text{Verify}(x, y)]$, all correct responses within a rollout group receive the same reward, so GRPO's group-relative advantage collapses to an identical value across all $K_c$ correct trajectories and cannot preferentially reinforce higher-quality reasoning.
As training progresses and $K_c \to K$, this further induces instabilities in the policy update.
GAR addresses both pathologies by introducing intra-group variance among correct responses through gradient-space alignment with expert reasoning traces.
An end-to-end overview is provided in Figure~\ref{fig:backbone}.

\subsection{The Gradient Alignment Hypothesis}
\label{sec:alignment-hypothesis}

Let $\mathcal{L}(\theta; x, y) = -\frac{1}{|y|}\sum_{t=1}^{|y|} \log \pi_\theta(y_t \mid x, y_{<t})$ denote the teacher-forcing negative log-likelihood of a response $y$ under the current policy.
Consider a prompt $x$ together with two correct responses: a candidate $y$ generated by the policy and an expert anchor $a$.
Define the (full-parameter) gradient operator
\begin{equation}
    \mathbf{g}(x, y) = \nabla_\theta \mathcal{L}(\theta; x, y) \in \mathbb{R}^{|\theta|}.
\label{eq:full-grad}
\end{equation}
The gradient alignment hypothesis posits that whenever the candidate response implements the same underlying reasoning process as the anchor, the two gradients are close in direction, i.e., $\cos(\mathbf{g}(x, y), \mathbf{g}(x, a)) \approx 1$, whereas trajectories that arrive at the correct answer through qualitatively different computational pathways yield gradients that are only weakly aligned.
The intuition is that a policy gradient step on either trajectory nudges the same set of reasoning-relevant computational circuits, whereas semantically incompatible trajectories modify non-overlapping subnetworks.

Computing the full gradient $\mathbf{g}(x, y) \in \mathbb{R}^{|\theta|}$ for every rollout is prohibitive in practice.
Our key algorithmic contribution is to replace the full gradient with a truncated surrogate obtained by freezing the Transformer body and backpropagating only through the output projection layer, resulting in a vector of dimension $d$ rather than $|\theta|$.
We show in Section~\ref{sec:theory} that the cosine similarity of these truncated surrogates is bi-Lipschitz equivalent to the normalized output-layer NTK similarity, preserving relative ordering among trajectories under mild regularity conditions.

\subsection{Truncated Gradient Signal}
\label{sec:gradient_signal}

For a prompt-response pair $(x, y)$, we perform a no-grad forward pass through the full Transformer and intercept the hidden state $\mathbf{H} \in \mathbb{R}^{n \times d}$ at the input of the output projection via a forward pre-hook, detach it to block gradient flow into the Transformer body, and compute a truncated cross-entropy loss through only the LM head:
\begin{equation}
    \mathcal{L} = \frac{1}{|\mathcal{T}_y|} \sum_{t \in \mathcal{T}_y} \ell\!\left(W_o \widetilde{\mathbf{h}}_t + \mathbf{b}_o,\; y_{t+1}\right),
\label{eq:ce_loss}
\end{equation}
where $\mathcal{T}_y$ indexes the response span (or, when delimited, the explicit thinking span), $\widetilde{\mathbf{h}}_t$ is the detached hidden state at position $t$, $y_{t+1}$ is the next-token target, and $W_o, \mathbf{b}_o$ are the output projection parameters.
The gradient $\mathbf{G}_t = \partial \mathcal{L} / \partial \widetilde{\mathbf{h}}_t$ is then combined element-wise with the activation to form the gradient-activation signal
\begin{equation}
    \mathbf{S}_t = \mathbf{G}_t \odot \widetilde{\mathbf{h}}_t, \quad \forall\, t \in \mathcal{T}_y,
\label{eq:grad-act}
\end{equation}
The element-wise product emphasizes dimensions where loss sensitivity and activation magnitude jointly concentrate, making the signal more robust to surface-level wording variation than the raw gradient alone.
Proposition~\ref{prop:grad-act-linear} shows that $\mathbf{S}_t$ admits a first-order interpretation as the linearized per-dimension contribution to the log-likelihood.
The per-token signals are averaged across response positions and L2-normalized,
\begin{equation}
    \mathbf{v} = \frac{\bar{\mathbf{s}}}{\|\bar{\mathbf{s}}\|_2}, \quad \bar{\mathbf{s}} = \frac{1}{|\mathcal{T}_y|} \sum_{t \in \mathcal{T}_y} \mathbf{S}_t,
\label{eq:normalize}
\end{equation}
which removes magnitude drift across training steps so that cosine comparisons remain commensurate across trajectories of different lengths.

\subsection{Reward Formulation}
\label{sec:reward_formulation}

\paragraph{Verifier Gate.}
GAR retains the outcome verifier as a hard gate: a response $y_i$ with binary outcome reward $r_\text{raw}(x, y_i) = 0$ receives no alignment bonus and triggers no gradient computation, which both prevents the policy from inflating its reward through spurious directions in gradient space and confines the additional cost of GAR to verifier-passing rollouts.

\paragraph{Anchor and Cosine Score.}
For each prompt $x$ we designate an expert anchor $a(x)$: the chain-of-thought solution shipped with the training dataset.
Crucially, widely used math corpora such as NuminaMath-CoT~\cite{li2024numinamath} already provide such solutions as standard metadata, yet conventional GRPO pipelines ignore them entirely.
GAR repurposes these annotations as gradient-space references, converting an underutilized resource into dense process supervision at no additional annotation cost.
The anchor is mapped to its gradient vector $\mathbf{v}_a$ via the same truncated procedure, with the anchor vector cached per prompt across the rollout batch to avoid redundant computation.
For a verifier-passing response $y_i$ with gradient vector $\mathbf{v}_i$, the alignment score is the cosine similarity with the anchor, which reduces to the inner product because both vectors are L2-normalized:
\begin{equation}
    b(y_i) = \mathbf{v}_i^\top \mathbf{v}_a.
\label{eq:best_cosine}
\end{equation}

\paragraph{Final Reward.}
The final GAR reward applies the gated, group-centered, non-negatively clipped bonus on top of the base reward:
\begin{equation}
\begin{aligned}
    &r_\text{GAR}(x, y_i) = \\
    &\begin{cases}
        r_\text{base} + \beta \cdot \max(0, \hat{b}(y_i)) + p(y_i), & r_\text{raw} > 0, \\
        p(y_i), & \text{otherwise},
    \end{cases}
\end{aligned}
\label{eq:final_reward}
\end{equation}
where $\hat{b}(y_i) = b(y_i) - \frac{1}{|\mathcal{P}(x)|}\sum_{y_j \in \mathcal{P}(x)} b(y_j)$ is the bonus centered over the correct subset $\mathcal{P}(x)$ of the rollout group, $r_\text{base}$ (default $1.0$) is the base reward for correctness, $\beta \ge 0$ (default $0.5$) scales the alignment bonus, and $p(y_i) \le 0$ is a small format penalty on responses that violate the prescribed think/answer structure.
The $\max(0, \cdot)$ clip keeps the GAR reward weakly above the outcome-only baseline on every correct trajectory and is essential for the safe reward-shaping guarantee of Theorem~\ref{thm:policy-invariance}.

\section{Theoretical Analysis}
\label{sec:theory}

All proofs are deferred to the appendix.

\subsection{NTK Interpretation and Multiplicative Decomposition}
\label{sec:theory-ntk}

Let $\mathbf{g}_o(x, y) = \nabla_{W_o} \mathcal{L}(W_o; x, y) \in \mathbb{R}^{|\mathcal{V}| \times d}$ denote the output-layer gradient.
Straightforward differentiation yields $\mathbf{g}_o(x, y) = \frac{1}{|\mathcal{T}_y|} \sum_{t} (\mathbf{p}_t - \mathbf{e}_{y_{t+1}}) \otimes \mathbf{h}_t$.

\begin{proposition}[Gradient-activation product as linearized log-likelihood]
\label{prop:grad-act-linear}
Let $\mathbf{H}_y \in \mathbb{R}^{n \times d}$ denote the matrix of final-layer hidden states at the input of the output projection, obtained from a no-gradient forward pass. The per-token signal $\mathbf{u}_t = \mathbf{G}_t \odot \mathbf{h}_t$ of Eq.~\eqref{eq:grad-act} satisfies $\frac{d}{d\eta} \mathcal{L}\big((1+\eta)\mathbf{H}_y\big)\big|_{\eta=0} = \frac{1}{|\mathcal{T}_y|} \sum_{t} \mathbf{1}^\top \mathbf{u}_t$, so the aggregated signal $\bar{\mathbf{s}}$ captures the linearized contribution of each hidden dimension to the log-likelihood.
\end{proposition}

\begin{theorem}[NTK-induced functional alignment]
\label{thm:ntk-bound}
Let $\Theta_o(y, a) = \langle \mathrm{vec}(\mathbf{g}_o(x, y)), \mathrm{vec}(\mathbf{g}_o(x, a)) \rangle$ denote the output-layer empirical NTK, and let $\mathbf{v}_y, \mathbf{v}_a$ denote the L2-normalized gradient-activation vectors of Eq.~\eqref{eq:normalize}.
Under bounded error signals ($\|\bar{\boldsymbol{\delta}}_y\|_2 \le B_\delta$) and hidden states ($\|\bar{\mathbf{h}}_y\|_2 \le B_h$), there exist constants $c_1, c_2 > 0$ such that
\begin{equation}
\begin{aligned}
    c_1 \cos(\mathbf{v}_y, \mathbf{v}_a) &\le \frac{\Theta_o(y, a)}{\|\mathbf{g}_o(x, y)\|_F \|\mathbf{g}_o(x, a)\|_F} \\
    &\le c_2 \cos(\mathbf{v}_y, \mathbf{v}_a) + \mathcal{O}(\kappa_y + \kappa_a),
\end{aligned}
\label{eq:ntk-bound}
\end{equation}
where $\kappa_y, \kappa_a$ measure the per-token error dispersion around the trajectory means.
\end{theorem}

Theorem~\ref{thm:ntk-bound} shows that rewarding high-cosine trajectories is equivalent, up to bounded distortion, to rewarding high output-layer NTK similarity with the expert anchor.

\begin{theorem}[Multiplicative decomposition of gradient alignment]
\label{thm:decomposition}
Under the same assumptions, the NTK inner product decomposes as
\begin{equation}
\begin{aligned}
    &\langle \mathbf{g}_o(x, y),\; \mathbf{g}_o(x, a) \rangle \\
    &= \underbrace{\big(\bar{\boldsymbol{\delta}}^{y\top} \bar{\boldsymbol{\delta}}^{a}\big)}_{\text{gradient-direction}}
       \underbrace{\big(\bar{\mathbf{h}}^{y\top} \bar{\mathbf{h}}^{a}\big)}_{\text{activation-pattern}} \\
    &\quad + \mathcal{O}\!\big(B_h^2 \kappa_y \kappa_a
       + B_\delta(\kappa_y + \kappa_a) B_h\big).
\end{aligned}
\label{eq:decomposition}
\end{equation}
\end{theorem}

The multiplicative structure requires agreement in \emph{both} prediction-error profile and activation pattern; a trajectory reaching the correct answer through a different computational pathway receives a low cosine score.
Empirically, across six training-set problems where correct rollouts use identifiably different solution methods, same-method rollouts receive $3.5\times$ higher cosine scores than alternative correct methods (mean $\bar{b}_{\text{same}} = 0.45$ vs.\ $\bar{b}_{\text{diff}} = 0.13$; Appendix~\ref{app:alignment-diversity}), confirming that the signal discriminates based on the underlying derivation strategy rather than surface-level correlates such as length or formatting.

\begin{corollary}[Orthogonality under subspace separation]
\label{cor:orthogonality}
If the $\varepsilon$-effective supports $\mathcal{S}_\varepsilon^y$ and $\mathcal{S}_\varepsilon^a$ are disjoint, then $|\cos(\mathbf{v}_y, \mathbf{v}_a)| \le 2\varepsilon d + \mathcal{O}(\kappa_y + \kappa_a)$.
\end{corollary}

\subsection{Safe Reward Shaping}
\label{sec:theory-invariance}

\begin{definition}[Outcome-verified optimal policy]
\label{def:opt-policy}
A policy $\pi^\star$ is outcome-verified optimal if it maximizes $\mathcal{J}_0(\pi) = \mathbb{E}_{x, y \sim \pi}\,\mathbf{1}[\mathrm{Verify}(x, y)]$.
\end{definition}

\begin{theorem}[Safe reward shaping]
\label{thm:policy-invariance}
The GAR reward of Eq.~\eqref{eq:final_reward} satisfies two properties.
(i)~\textbf{Non-degradation}: for every correct response $y$ with $r_\text{raw}(x,y)>0$, $r_\text{GAR}(x,y) - p(y) \ge r_\text{base} > 0$, so GAR never reduces the reward of a correct response below the outcome-only baseline.
(ii)~\textbf{Strict incentive separation}: for every incorrect response $y'$ with $r_\text{raw}(x,y')=0$, $r_\text{GAR}(x,y') = p(y') \le 0 < r_\text{base} \le r_\text{GAR}(x,y) - p(y)$, so the policy gradient consistently assigns higher advantage to correct responses than to incorrect ones.
\end{theorem}

Direction~(i) follows from the verifier gate and $\max(0, \hat{b})$ clip ensuring every correct response contributes at least $r_\text{base}$; direction~(ii) follows from the verifier gate assigning $r_\text{GAR}(x,y') = p(y') \le 0$ to incorrect responses, while every correct response receives at least $r_\text{base} > 0$ on top of the format penalty.

\subsection{Variance and Unbiasedness}
\label{sec:theory-variance}

\begin{proposition}[Variance amplification]
\label{prop:variance}
Under the flat outcome reward, $\mathrm{Var}_{i : r_\text{raw}=1}[A_i^\text{outcome}] = 0$.
Under GAR, $\mathrm{Var}_{i : r_\text{raw}=1}[A_i^\text{GAR}] > 0$ whenever the alignment bonuses have non-zero variance within the correct subset.
\end{proposition}

\begin{proposition}[Unbiasedness of prompt-group normalization]
\label{prop:unbiased}
The group-normalized bonus satisfies $\mathbb{E}_{y : r_\text{raw}=1}[\hat{b}(y)] = 0$ for every prompt $x$.
\end{proposition}

\subsection{Monotonic Alignment Improvement}
\label{sec:theory-monotonic}

\begin{proposition}[Monotonic alignment improvement]
\label{prop:monotonic}
Let $\bar{C}_t = \mathbb{E}_{x, y \sim \pi_t, r_\text{raw}=1}[\cos(\mathbf{v}_y, \mathbf{v}_a)]$ denote the expected cosine alignment at iteration $t$.
Under GAR-shaped GRPO updates with sufficiently small KL penalty, $\bar{C}_{t+1} \ge \bar{C}_t$.
\end{proposition}

This follows because GAR assigns positive advantages exclusively to above-mean-cosine trajectories, so $\mathrm{Cov}[A_i, c_i] \ge 0$.

\section{Experimental Setup}
\label{sec:experiments}
\begin{table*}[!t]
\centering
\small
\vspace{-2mm}
\caption{Pass@$k$ accuracy (\%, abbreviated as P@$k$ in headers) on competition-level math benchmarks, averaged over 10 independent runs (standard deviations in {\color{gray}gray}; all GAR improvements over the corresponding base optimizer are statistically significant with $p < 0.05$ by paired $t$-test across runs). All methods train from Qwen3 base checkpoints without SFT warmup using identical training data and compute budget. Best results per model in \textbf{bold}.}
\vspace{-2mm}
\label{tab:main_results}
\begin{adjustbox}{max width=\textwidth}
\begin{tabular}{l ccc ccc ccc ccc}
\toprule
\rowcolor{lightpurple} & \multicolumn{3}{c}{\textbf{IMO-AnswerBench}} & \multicolumn{3}{c}{\textbf{HMMT '26}} & \multicolumn{3}{c}{\textbf{HMMT '25}} & \multicolumn{3}{c}{\textbf{AIME '26}} \\
\cmidrule(lr){2-4} \cmidrule(lr){5-7} \cmidrule(lr){8-10} \cmidrule(lr){11-13}
\rowcolor{lightpurple} \textbf{Method} & P@1 & P@4 & P@16 & P@1 & P@4 & P@16 & P@1 & P@4 & P@16 & P@1 & P@4 & P@16 \\
\midrule
\rowcolor{skyblue} \multicolumn{13}{c}{\textit{Qwen3-4B-Base}} \\
Base (no RL)                    & 3.05\sd{0.9} & 9.72\sd{1.1} & 21.91\sd{1.0} & 0.61\sd{1.0} & 2.62\sd{1.1} & 5.76\sd{2.2} & 0.17\sd{1.3} & 1.36\sd{1.0} & 4.34\sd{1.7} & 1.17\sd{1.2} & 4.85\sd{1.9} & 13.00\sd{2.7} \\
+ REINFORCE++                   & 6.86\sd{0.9} & 16.22\sd{1.0} & 29.76\sd{1.7} & 3.33\sd{1.4} & 9.01\sd{1.6} & 18.03\sd{1.8} & 3.17\sd{0.8} & 7.18\sd{1.5} & 15.33\sd{2.0} & 7.67\sd{0.9} & 17.86\sd{1.4} & 29.34\sd{2.1} \\
+ GRPO                          & 7.96\sd{0.7} & 17.91\sd{1.2} & 31.72\sd{1.3} & 2.42\sd{1.1} & 7.90\sd{1.6} & 15.30\sd{1.6} & 3.83\sd{0.8} & 8.32\sd{1.1} & 13.51\sd{2.0} & 6.83\sd{1.3} & 16.81\sd{2.1} & 27.67\sd{2.5} \\
+ MASPO                         & 7.01\sd{0.8} & 16.21\sd{1.0} & 30.32\sd{1.5} & 2.73\sd{1.2} & 8.27\sd{1.4} & 15.76\sd{2.4} & 3.67\sd{1.0} & 8.37\sd{1.2} & 16.49\sd{2.2} & 7.83\sd{1.3} & 18.28\sd{2.1} & 30.33\sd{2.4} \\
+ Grad2Reward                   & 7.76\sd{0.7} & 17.56\sd{0.9} & 31.46\sd{1.2} & 2.58\sd{1.0} & 7.85\sd{1.2} & 15.60\sd{2.2} & 3.33\sd{1.0} & 7.35\sd{1.2} & 13.17\sd{2.3} & 6.50\sd{1.5} & 17.09\sd{2.0} & 30.50\sd{2.8} \\
+ G2RL                          & 8.10\sd{0.7} & 18.27\sd{1.0} & 32.59\sd{1.4} & 2.88\sd{1.3} & 8.24\sd{1.5} & 16.06\sd{2.3} & 3.50\sd{0.9} & 7.51\sd{1.1} & 13.50\sd{2.1} & 7.33\sd{1.1} & 17.66\sd{1.6} & 30.67\sd{2.9} \\
\rowcolor{lightorange} + GAR-REINFORCE++               & 7.05\sd{0.7} & 16.57\sd{0.9} & 30.88\sd{1.1} & \textbf{4.39}\sd{0.9} & \textbf{10.70}\sd{1.2} & \textbf{19.24}\sd{2.3} & 4.83\sd{1.0} & \textbf{9.79}\sd{1.4} & \textbf{17.00}\sd{2.0} & 8.00\sd{1.1} & \textbf{18.39}\sd{1.7} & \textbf{31.50}\sd{2.5} \\
\rowcolor{lightorange} \quad\textit{Improv.}           & \textcolor{crimson}{+2.8\%} & \textcolor{crimson}{+2.2\%} & \textcolor{crimson}{+3.8\%} & \textcolor{crimson}{+31.8\%} & \textcolor{crimson}{+18.8\%} & \textcolor{crimson}{+6.7\%} & \textcolor{crimson}{+52.4\%} & \textcolor{crimson}{+36.4\%} & \textcolor{crimson}{+10.9\%} & \textcolor{crimson}{+4.3\%} & \textcolor{crimson}{+3.0\%} & \textcolor{crimson}{+7.4\%} \\
\rowcolor{lightorange} + GAR-GRPO                      & \textbf{8.76}\sd{0.6} & \textbf{19.06}\sd{1.2} & \textbf{33.76}\sd{1.2} & 3.18\sd{1.2} & 9.00\sd{1.6} & 16.52\sd{1.8} & \textbf{5.00}\sd{0.8} & 9.19\sd{1.1} & 15.50\sd{2.0} & \textbf{8.50}\sd{0.9} & 18.05\sd{1.6} & 29.66\sd{3.0} \\
\rowcolor{lightorange} \quad\textit{Improv.}           & \textcolor{crimson}{+10.1\%} & \textcolor{crimson}{+6.4\%} & \textcolor{crimson}{+6.4\%} & \textcolor{crimson}{+31.4\%} & \textcolor{crimson}{+13.9\%} & \textcolor{crimson}{+8.0\%} & \textcolor{crimson}{+30.5\%} & \textcolor{crimson}{+10.5\%} & \textcolor{crimson}{+14.7\%} & \textcolor{crimson}{+24.5\%} & \textcolor{crimson}{+7.4\%} & \textcolor{crimson}{+7.2\%} \\
\midrule
\rowcolor{skyblue} \multicolumn{13}{c}{\textit{Qwen3-8B-Base}} \\
Base (no RL)                    & 3.54\sd{0.8} & 11.03\sd{0.9} & 23.79\sd{1.2} & 0.91\sd{1.1} & 2.69\sd{1.2} & 7.57\sd{2.4} & 0.67\sd{1.2} & 1.81\sd{1.2} & 6.49\sd{2.4} & 2.00\sd{1.0} & 5.84\sd{2.0} & 12.16\sd{2.8} \\
+ REINFORCE++                   & 6.83\sd{0.5} & 16.27\sd{0.8} & 29.74\sd{1.3} & 4.09\sd{1.0} & 8.12\sd{1.5} & 14.85\sd{2.3} & 4.33\sd{0.9} & 7.50\sd{1.6} & 19.84\sd{1.7} & 8.17\sd{1.3} & 18.44\sd{1.9} & 33.00\sd{3.0} \\
+ GRPO                          & 6.99\sd{0.6} & 16.50\sd{1.2} & 31.05\sd{1.7} & 4.24\sd{1.4} & 9.38\sd{1.6} & 15.46\sd{1.9} & 4.50\sd{1.0} & 8.01\sd{1.4} & 18.00\sd{2.1} & 8.00\sd{1.3} & 18.79\sd{1.6} & 33.83\sd{3.2} \\
+ MASPO                         & 6.60\sd{0.8} & 15.70\sd{0.7} & 30.89\sd{1.6} & 4.70\sd{1.0} & 8.71\sd{1.2} & 15.00\sd{1.7} & 4.50\sd{1.0} & 9.51\sd{1.1} & 19.67\sd{1.9} & 8.50\sd{1.4} & 19.17\sd{1.7} & 34.17\sd{3.1} \\
+ Grad2Reward                   & 6.24\sd{0.7} & 15.29\sd{0.9} & 29.76\sd{1.0} & 4.09\sd{0.9} & 7.77\sd{1.5} & 13.94\sd{2.3} & 3.67\sd{0.9} & 8.82\sd{1.2} & 17.84\sd{2.0} & 7.17\sd{1.4} & 17.49\sd{1.5} & 34.66\sd{3.2} \\
+ G2RL                          & 7.53\sd{0.8} & 17.31\sd{1.0} & 31.93\sd{1.0} & 4.39\sd{0.8} & 9.88\sd{1.8} & 16.21\sd{2.5} & 4.83\sd{1.0} & 8.63\sd{1.6} & 18.83\sd{1.8} & 8.33\sd{1.1} & 19.34\sd{1.6} & 35.00\sd{2.9} \\
\rowcolor{lightorange} + GAR-REINFORCE++               & 7.76\sd{0.8} & \textbf{18.14}\sd{0.7} & 32.38\sd{1.0} & 4.85\sd{1.0} & 9.78\sd{1.4} & \textbf{17.88}\sd{2.4} & 5.83\sd{1.2} & 10.04\sd{1.4} & \textbf{21.84}\sd{2.3} & 8.67\sd{1.0} & 18.77\sd{1.8} & 35.50\sd{2.9} \\
\rowcolor{lightorange} \quad\textit{Improv.}           & \textcolor{crimson}{+13.6\%} & \textcolor{crimson}{+11.5\%} & \textcolor{crimson}{+8.9\%} & \textcolor{crimson}{+18.6\%} & \textcolor{crimson}{+20.4\%} & \textcolor{crimson}{+20.4\%} & \textcolor{crimson}{+34.6\%} & \textcolor{crimson}{+33.9\%} & \textcolor{crimson}{+10.1\%} & \textcolor{crimson}{+6.1\%} & \textcolor{crimson}{+1.8\%} & \textcolor{crimson}{+7.6\%} \\
\rowcolor{lightorange} + GAR-GRPO                      & \textbf{8.14}\sd{0.5} & 18.13\sd{1.1} & \textbf{33.73}\sd{1.3} & \textbf{5.15}\sd{1.0} & \textbf{10.62}\sd{1.3} & 17.73\sd{2.5} & \textbf{6.00}\sd{1.0} & \textbf{10.83}\sd{1.3} & 19.83\sd{2.3} & \textbf{9.17}\sd{1.5} & \textbf{19.95}\sd{2.0} & \textbf{36.49}\sd{2.4} \\
\rowcolor{lightorange} \quad\textit{Improv.}           & \textcolor{crimson}{+16.5\%} & \textcolor{crimson}{+9.9\%} & \textcolor{crimson}{+8.6\%} & \textcolor{crimson}{+21.5\%} & \textcolor{crimson}{+13.2\%} & \textcolor{crimson}{+14.7\%} & \textcolor{crimson}{+33.3\%} & \textcolor{crimson}{+35.2\%} & \textcolor{crimson}{+10.2\%} & \textcolor{crimson}{+14.6\%} & \textcolor{crimson}{+6.2\%} & \textcolor{crimson}{+7.9\%} \\
\bottomrule
\end{tabular}
\end{adjustbox}
\end{table*}

\subsection{Training}
\label{sec:setup}

GAR is implemented within the SLIME~/~Megatron training stack as an online reward hook invoked before GRPO advantage computation (Appendix~\ref{app:impl}).
We train Qwen3-4B-Base and Qwen3-8B-Base~\cite{yang2025qwen3} from their base checkpoints without SFT warmup using full-parameter reinforcement learning, for $400$ steps with batch size $128$ and $K{=}16$ rollouts per prompt on ${\sim}$10k NuminaMath-CoT~\cite{li2024numinamath} problems; each problem's chain-of-thought solution serves as the expert anchor.
Each optimization step therefore contains $2{,}048$ sampled trajectories before verifier gating, which provides multiple candidate solutions per prompt for estimating within-group reward differences.
Training directly from the base checkpoints prevents the observed gains from being attributed to an SFT warmup and isolates the contribution of the reward signal.
The default GAR hyperparameters are $\beta{=}0.5$, max GAR span $L{=}768$ tokens, activation threshold $\tau_a{=}0.05$, and anchor filter $p_f{=}0.7$.
The expert-anchor vector is cached within each rollout group, and the alignment bonus is evaluated only for verifier-passing responses, limiting additional computation to the trajectories for which process-level differentiation is meaningful.

\subsection{Baselines}
\label{sec:baselines}

We compare against \textbf{GRPO}~\cite{shao2024deepseekmath,yu2025dapo}, \textbf{REINFORCE++}~\cite{hu2025reinforce++}, \textbf{MASPO}~\cite{fu2026maspo}, \textbf{Grad2Reward}~\cite{zhang2026grad2reward}, and \textbf{G2RL}~\cite{liang2025can}.
We report \textbf{GAR-GRPO} and \textbf{GAR-REINFORCE++} to isolate GAR's contribution from the optimizer.
GRPO and REINFORCE++ provide two distinct policy optimization backbones, while MASPO, Grad2Reward, and G2RL test whether GAR remains competitive with recent methods designed to improve signal reliability or exploit gradient information.
All methods share the same base checkpoint, training data, compute budget, and rollout infrastructure; only the reward function differs.
Consequently, each comparison between an optimizer and its GAR variant measures the effect of reward shaping without conflating it with additional data, supervised initialization, or a larger rollout budget.

\subsection{Evaluation}
\label{sec:eval}

We report pass@$k$ ($k{\in}\{1,4,16\}$) on four held-out math benchmarks (IMO-AnswerBench, HMMT~'25/\!'26, AIME~'26) and two general reasoning benchmarks (GPQA Diamond, MMLU-Pro) to assess cross-domain transfer.
Correctness is determined by exact answer match after normalization.
Pass@1 measures whether training concentrates probability mass on a correct solution in a single attempt, whereas pass@16 measures whether the policy preserves broader solution coverage across repeated samples.
We average every configuration over 10 independently trained runs and use paired significance tests under matched sampling conditions; complete decoding, statistical, and contamination-control protocols are provided in Appendix~\ref{app:eval}.

\section{Results}
\label{sec:results}

\subsection{Main Results}
\label{sec:main_results}

Table~\ref{tab:main_results} presents the main results across four competition-level benchmarks.
GAR improves over the corresponding base optimizer on every benchmark--model combination, with relative gains of up to 52.4\% at pass@1 (HMMT~2025, 4B).
The improvements are most pronounced at lower $k$, indicating that GAR steers the policy toward higher-probability correct solutions rather than merely expanding coverage.
For GAR-GRPO at the 4B scale, pass@1 increases from 2.42 to 3.18 on HMMT~2026 and from 3.83 to 5.00 on HMMT~2025, corresponding to relative gains of 31.4\% and 30.5\%, respectively.
The same comparisons at the 8B scale increase from 4.24 to 5.15 and from 4.50 to 6.00, showing that the benefit persists as model capacity grows.
Improvements also remain positive at pass@16 across every dataset and optimizer, which indicates that concentrating probability on strong solutions does not reduce the overall coverage of correct reasoning paths.
Layering GAR on REINFORCE++~\cite{hu2025reinforce++} yields comparable gains to the GRPO variant, demonstrating that the reward signal is complementary to optimizer-side design choices.
The largest relative gain occurs for GAR-REINFORCE++ on HMMT~2025 at the 4B scale, where pass@1 rises from 3.17 to 4.83.
Among gradient-based competitors, Grad2Reward~\cite{zhang2026grad2reward} and G2RL~\cite{liang2025can} underperform GAR across all benchmarks, while MASPO~\cite{fu2026maspo} provides only modest improvement over GRPO.
Together, these results indicate that the principal advantage arises from aligning the reward with expert reasoning in gradient space rather than from the choice of policy optimizer alone.

\section{Empirical Analysis}
\label{sec:analysis}

\subsection{General Reasoning Transfer}
\label{sec:general_reasoning}

\begin{table}[t]
\centering
\small
\caption{General reasoning transfer (Qwen3-4B-Base, 10 runs). All methods are trained exclusively on mathematical data and evaluated zero-shot. GPQA reports pass@$k$ and maj@16; MMLU-Pro reports micro-averaged pass@1.}
\label{tab:general_reasoning}
\begin{adjustbox}{max width=0.9\columnwidth}
\begin{tabular}{l ccc c}
\toprule
\rowcolor{lightpurple} & \multicolumn{3}{c}{\textbf{GPQA Diamond}} & \textbf{MMLU-Pro} \\
\cmidrule(lr){2-4} \cmidrule(lr){5-5}
\rowcolor{lightpurple} \textbf{Method} & P@1 & P@4 & Maj@16 & Avg.\ P@1 \\
\midrule
GRPO          & 28.31 & 47.73 & 41.52 & 48.17 \\
MASPO         & 29.82 & 48.94 & 42.68 & 48.24 \\
G2RL          & 30.20 & 47.25 & 41.06 & 47.13 \\
\rowcolor{lightorange} GAR-GRPO      & \textbf{30.88} & \textbf{49.52} & \textbf{43.46} & \textbf{50.58} \\
\bottomrule
\end{tabular}
\end{adjustbox}
\end{table}

To assess whether gradient-aligned rewards transfer beyond the training domain, Table~\ref{tab:general_reasoning} evaluates all methods, trained exclusively on mathematical data, on two general reasoning benchmarks.
GAR-GRPO outperforms all baselines on both GPQA Diamond and MMLU-Pro, indicating that the gradient alignment signal captures domain-general reasoning structure rather than math-specific heuristics.
Relative to GRPO, the 4B model gains 2.57 points on GPQA pass@1, 1.79 points on pass@4, 1.94 points on majority voting, and 2.41 points on MMLU-Pro.
GAR-GRPO also exceeds the strongest alternative baseline in every reported column, so the transfer improvement is not explained solely by a weak GRPO reference.
Results on Qwen3-8B-Base (Appendix~\ref{app:general-8b}) confirm consistent improvements at the larger scale.
At 8B, the absolute gains over GRPO remain between 2.02 and 2.29 points across all four metrics, indicating stable transfer across model scales.

\subsection{Ablation Study}
\label{sec:ablation}

\begin{figure}[h]
\centering
\includegraphics[width=\columnwidth]{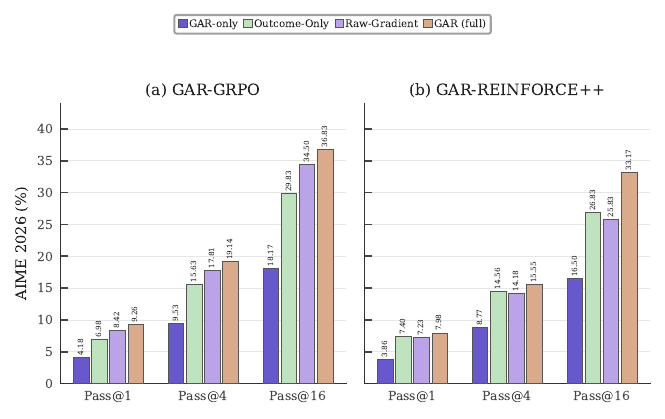}
\caption{Ablation on Qwen3-8B-Base (AIME~2026), removing one GAR component at a time from (a)~GAR-GRPO and (b)~GAR-REINFORCE++. Activation weighting is essential for consistent gains across both optimizers.}
\label{fig:ablation}
\end{figure}

Figure~\ref{fig:ablation} isolates the contribution of each GAR component on Qwen3-8B across two optimizers.
Full GAR consistently achieves the best performance regardless of optimizer choice.
Dropping the outcome reward entirely (GAR-only) degrades performance well below the baseline, confirming that the verifier gate is essential.
This result shows that gradient alignment is effective as a process-level refinement of correctness but is not a substitute for the binary outcome signal.
Interestingly, using raw gradients without activation weighting yields inconsistent results: it helps with GRPO but slightly hurts with REINFORCE++, suggesting that the unweighted gradient signal is noisy and optimizer-sensitive.
The activation-weighted formulation of Eq.~\eqref{eq:grad-act} resolves this instability, delivering reliable gains in both settings.
The agreement across both optimizer backbones further indicates that activation weighting is a structural component of the reward rather than an optimizer-specific tuning effect.

\subsection{Reward Distribution}
\label{sec:reward_dist}

\begin{figure}[t]
\centering
\includegraphics[width=0.8\columnwidth]{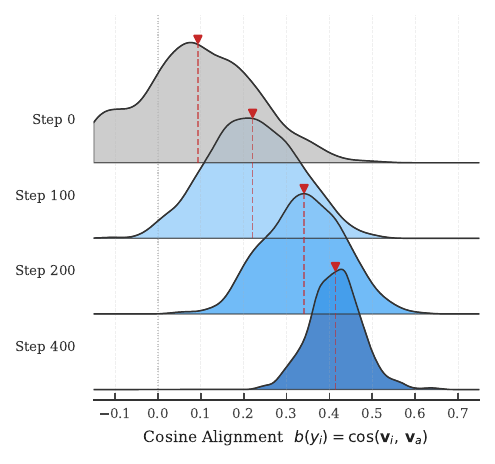}
\caption{Cosine alignment distribution among correct rollouts at training steps 0, 100, 200, 400 (Qwen3-8B). Red dashed lines mark means.}
\label{fig:reward_dist}
\end{figure}

Figure~\ref{fig:reward_dist} shows how the cosine alignment distribution evolves during training.
The mean $b(y_i)$ increases from 0.10 to 0.42 while variance narrows, confirming progressive alignment with expert anchors.
Because the distribution contains only verifier-passing rollouts, the fourfold increase cannot be attributed merely to a higher fraction of correct answers.
Instead, it reflects a redistribution within the correct subset toward trajectories whose gradient directions more closely match the expert solution.
The simultaneous reduction in variance suggests that this behavior becomes systematic across rollouts rather than being driven by a small number of highly aligned outliers.

\subsection{Computational Overhead}
\label{sec:overhead}

\begin{figure}[t]
\centering
\includegraphics[width=0.75\columnwidth]{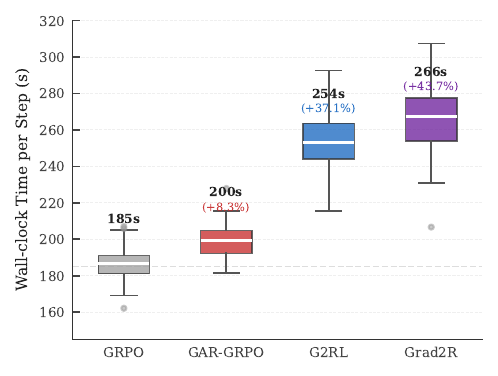}
\caption{Per-step wall-clock time distribution (Qwen3-8B, 4$\times$A100, $K{=}16$, 100 steps). GAR adds only 8.3\% overhead via \texttt{lm\_head}-only backpropagation.}
\label{fig:overhead}
\end{figure}

Figure~\ref{fig:overhead} compares per-step training cost across gradient-based methods.
GAR's truncated backpropagation through only the \texttt{lm\_head} layer adds just 8.3\% overhead over outcome-only GRPO.
G2RL's pairwise gradient diversity computation requires multi-layer backpropagation (+37.1\%), while Grad2Reward~\cite{zhang2026grad2reward} performs a full backward pass through the Judge for each rubric (+43.7\%).
Thus, GAR incurs less than one quarter of G2RL's additional cost and less than one fifth of Grad2Reward's additional cost under the reported setup.
This efficiency is consistent with the design of GAR: gradients are truncated at the output projection layer, incorrect responses are removed by the verifier gate before gradient extraction, and each expert anchor is cached across the rollout group.
The measured overhead also falls within the analytical range derived in Appendix~\ref{app:overhead-analysis}, connecting the implementation-level cost model with observed wall-clock behavior.

\subsection{Additional Analysis}
\label{sec:additional}

We provide further empirical analysis in the appendix: hyperparameter sensitivity (Appendix~\ref{sec:sensitivity}), analysis of alignment and solution diversity (Appendix~\ref{app:alignment-diversity}), and Qwen3-8B general reasoning transfer results (Appendix~\ref{app:general-8b}).

\section{Related Work}
\label{sec:related}

\paragraph{RL for LLM reasoning and reward design.}
Critic-free policy gradient methods~\cite{shao2024deepseekmath,yu2025dapo,fu2026maspo,hu2025reinforce++,liu2025understanding,melo2025stabilizing} and direct RL from base models~\cite{guo2025deepseek,zeng2025simplerl,wen2025light,aggarwal2025l1,yue2025vapo} have shown that chain-of-thought reasoning~\cite{wei2022chain} emerges from outcome rewards alone.
Process reward models~\cite{lightman2023let,uesato2022solving,wang2024math,luo2024improve,cui2025process,khalifa2025process,zhao2026genprm,wang2026grpo,cobbe2021training,snell2024scaling} offer denser supervision but require step-level annotations or a separately trained verifier.
Classical potential-based shaping~\cite{ng1999policy,harutyunyan2015expressing,devlin2012dynamic} establishes when shaped rewards preserve the optimal policy yet leaves the potential function unspecified.

\paragraph{Gradient signals in LLM training.}
Gradient information has been used for data attribution~\cite{koh2017understanding,pruthi2020estimating,park2023trak}, curriculum learning~\cite{mindermann2022prioritized,fifty2021efficiently}, and representation engineering~\cite{zou2023representation}, with NTK theory~\cite{jacot2018neural,mohamadi2023fast,tomihari2026learning} providing formal grounding.
Grad2Reward~\cite{zhang2026grad2reward}, G2RL~\cite{liang2025can}, and GradAlign~\cite{yang2026gradalign} recently apply gradient signals to LLM reward design, exploration diversity, and data selection, respectively.
GAR differs from these by anchoring the gradient signal to expert chain-of-thought solutions, combining the information richness of reference-based methods with the structural depth of gradient-space operation, while requiring no external judge or offline annotations.
Unlike recent distillation approaches that study information transfer and on-policy optimization granularity~\cite{fang2026distillationresistantlargelanguagemodels,li2026filterreweightrethinkingoptimization}, GAR does not optimize toward a fixed reference distribution: it preserves the RL exploration loop and uses expert CoTs only as a gradient-space reference signal.

\section{Conclusion}
\label{sec:conclusion}

We have presented Gradient-Aligned Reward (GAR), which converts readily available expert chain-of-thought solutions into gradient-space reference vectors and scores rollouts by cosine similarity with these anchors, providing dense process supervision within standard RLVR training at less than 9\% wall-clock overhead.
Experiments on four competition-level math benchmarks with Qwen3-4B and 8B base models show consistent pass@$k$ gains over GRPO, REINFORCE++, and gradient-based competitors, with positive transfer to GPQA Diamond and MMLU-Pro.

\section*{Acknowledgments}
This work is supported in part by the National Natural Science Foundation of China (No.~62372264 and No.~92467203 ). Chaokun Wang is the corresponding author.

\section*{Limitations}

While GAR demonstrates consistent improvements across two model scales and multiple benchmarks, it has not yet been deployed in an industrial production environment.

\section*{Ethics Statement}

GAR operates exclusively on publicly available mathematical reasoning benchmarks with open licenses, and does not involve human subjects, private data, or dual-use applications.

\bibliography{custom}

@article{guo2025deepseek,
  title={Deepseek-r1: Incentivizing reasoning capability in llms via reinforcement learning},
  author={Guo, Daya and Yang, Dejian and Zhang, Haowei and Song, Junxiao and Wang, Peiyi and Zhu, Qihao and Xu, Runxin and Zhang, Ruoyu and Ma, Shirong and Bi, Xiao and others},
  journal={arXiv preprint arXiv:2501.12948},
  year={2025}
}

@article{shao2024deepseekmath,
  title={Deepseekmath: Pushing the limits of mathematical reasoning in open language models},
  author={Shao, Zhihong and Wang, Peiyi and Zhu, Qihao and Xu, Runxin and Song, Junxiao and Bi, Xiao and Zhang, Haowei and Zhang, Mingchuan and Li, YK and Wu, Yang and others},
  journal={arXiv preprint arXiv:2402.03300},
  year={2024}
}

@inproceedings{lightman2023let,
  title={Let's verify step by step},
  author={Lightman, Hunter and Kosaraju, Vineet and Burda, Yuri and Edwards, Harrison and Baker, Bowen and Lee, Teddy and Leike, Jan and Schulman, John and Sutskever, Ilya and Cobbe, Karl},
  booktitle={The twelfth international conference on learning representations},
  year={2023}
}

@inproceedings{wang2024math,
  title={Math-shepherd: Verify and reinforce llms step-by-step without human annotations},
  author={Wang, Peiyi and Li, Lei and Shao, Zhihong and Xu, Runxin and Dai, Damai and Li, Yifei and Chen, Deli and Wu, Yu and Sui, Zhifang},
  booktitle={Proceedings of the 62nd Annual Meeting of the Association for Computational Linguistics (Volume 1: Long Papers)},
  pages={9426--9439},
  year={2024}
}

@article{luo2024improve,
  title={Improve mathematical reasoning in language models by automated process supervision},
  author={Luo, Liangchen and Liu, Yinxiao and Liu, Rosanne and Phatale, Samrat and Guo, Meiqi and Lara, Harsh and Li, Yunxuan and Shu, Lei and Zhu, Yun and Meng, Lei and others},
  journal={arXiv preprint arXiv:2406.06592},
  year={2024}
}

@inproceedings{wen2025light,
  title={Light-r1: Curriculum sft, dpo and rl for long cot from scratch and beyond},
  author={Wen, Liang and Cai, Yunke and Xiao, Fenrui and He, Xin and An, Qi and Duan, Zhenyu and Du, Yimin and Liu, Junchen and Tanglifu, Tanglifu and Lv, Xiaowei and others},
  booktitle={Proceedings of the 63rd Annual Meeting of the Association for Computational Linguistics (Volume 6: Industry Track)},
  pages={318--327},
  year={2025}
}

@article{li2024numinamath,
  title={Numinamath: The largest public dataset in ai4maths with 860k pairs of competition math problems and solutions},
  author={Li, Jia and Beeching, Edward and Tunstall, Lewis and Lipkin, Ben and Soletskyi, Roman and Huang, Shengyi and Rasul, Kashif and Yu, Longhui and Jiang, Albert Q and Shen, Ziju and others},
  journal={Hugging Face repository},
  volume={13},
  number={9},
  pages={9},
  year={2024}
}

@article{yang2025qwen3,
  title={Qwen3 technical report},
  author={Yang, An and Li, Anfeng and Yang, Baosong and Zhang, Beichen and Hui, Binyuan and Zheng, Bo and Yu, Bowen and Gao, Chang and Huang, Chengen and Lv, Chenxu and others},
  journal={arXiv preprint arXiv:2505.09388},
  year={2025}
}

@inproceedings{ng1999policy,
  title={Policy invariance under reward transformations: Theory and application to reward shaping},
  author={Ng, Andrew Y and Harada, Daishi and Russell, Stuart},
  booktitle={Icml},
  volume={99},
  pages={278--287},
  year={1999},
  organization={Citeseer}
}

@inproceedings{harutyunyan2015expressing,
  title={Expressing arbitrary reward functions as potential-based advice},
  author={Harutyunyan, Anna and Devlin, Sam and Vrancx, Peter and Now{\'e}, Ann},
  booktitle={Proceedings of the AAAI conference on artificial intelligence},
  volume={29},
  number={1},
  year={2015}
}

@inproceedings{devlin2012dynamic,
  title={Dynamic potential-based reward shaping},
  author={Devlin, Sam Michael and Kudenko, Daniel},
  booktitle={11th International Conference on Autonomous Agents and Multiagent Systems (AAMAS 2012)},
  pages={433--440},
  year={2012},
  organization={IFAAMAS}
}

@article{yu2025dapo,
  title={Dapo: An open-source llm reinforcement learning system at scale},
  author={Yu, Qiying and Zhang, Zheng and Zhu, Ruofei and Yuan, Yufeng and Zuo, Xiaochen and Yue, Yu and Dai, Weinan and Fan, Tiantian and Liu, Gaohong and Liu, Lingjun and others},
  journal={arXiv preprint arXiv:2503.14476},
  year={2025}
}

@article{aggarwal2025l1,
  title={L1: Controlling how long a reasoning model thinks with reinforcement learning},
  author={Aggarwal, Pranjal and Welleck, Sean},
  journal={arXiv preprint arXiv:2503.04697},
  year={2025}
}

@article{zeng2025simplerl,
  title={Simplerl-zoo: Investigating and taming zero reinforcement learning for open base models in the wild},
  author={Zeng, Weihao and Huang, Yuzhen and Liu, Qian and Liu, Wei and He, Keqing and Ma, Zejun and He, Junxian},
  journal={arXiv preprint arXiv:2503.18892},
  year={2025}
}

@article{cui2025process,
  title={Process reinforcement through implicit rewards},
  author={Cui, Ganqu and Yuan, Lifan and Wang, Zefan and Wang, Hanbin and Zhang, Yuchen and Chen, Jiacheng and Li, Wendi and He, Bingxiang and Fan, Yuchen and Yu, Tianyu and others},
  journal={arXiv preprint arXiv:2502.01456},
  year={2025}
}

@article{khalifa2025process,
  title={Process reward models that think},
  author={Khalifa, Muhammad and Agarwal, Rishabh and Logeswaran, Lajanugen and Kim, Jaekyeom and Peng, Hao and Lee, Moontae and Lee, Honglak and Wang, Lu},
  journal={arXiv preprint arXiv:2504.16828},
  year={2025}
}

@inproceedings{zhao2026genprm,
  title={Genprm: Scaling test-time compute of process reward models via generative reasoning},
  author={Zhao, Jian and Liu, Runze and Zhang, Kaiyan and Zhou, Zhimu and Gao, Junqi and Li, Dong and Lyu, Jiafei and Qian, Zhouyi and Qi, Biqing and Li, Xiu and others},
  booktitle={Proceedings of the AAAI Conference on Artificial Intelligence},
  volume={40},
  number={41},
  pages={34932--34940},
  year={2026}
}

@article{jacot2018neural,
  title={Neural tangent kernel: Convergence and generalization in neural networks},
  author={Jacot, Arthur and Gabriel, Franck and Hongler, Cl{\'e}ment},
  journal={Advances in neural information processing systems},
  volume={31},
  year={2018}
}

@inproceedings{koh2017understanding,
  title={Understanding black-box predictions via influence functions},
  author={Koh, Pang Wei and Liang, Percy},
  booktitle={International conference on machine learning},
  pages={1885--1894},
  year={2017},
  organization={PMLR}
}

@article{park2023trak,
  title={Trak: Attributing model behavior at scale},
  author={Park, Sung Min and Georgiev, Kristian and Ilyas, Andrew and Leclerc, Guillaume and Madry, Aleksander},
  journal={arXiv preprint arXiv:2303.14186},
  year={2023}
}

@article{fifty2021efficiently,
  title={Efficiently identifying task groupings for multi-task learning},
  author={Fifty, Chris and Amid, Ehsan and Zhao, Zhe and Yu, Tianhe and Anil, Rohan and Finn, Chelsea},
  journal={Advances in Neural Information Processing Systems},
  volume={34},
  pages={27503--27516},
  year={2021}
}

@inproceedings{mindermann2022prioritized,
  title={Prioritized training on points that are learnable, worth learning, and not yet learnt},
  author={Mindermann, S{\"o}ren and Brauner, Jan M and Razzak, Muhammed T and Sharma, Mrinank and Kirsch, Andreas and Xu, Winnie and H{\"o}ltgen, Benedikt and Gomez, Aidan N and Morisot, Adrien and Farquhar, Sebastian and others},
  booktitle={International Conference on Machine Learning},
  pages={15630--15649},
  year={2022},
  organization={PMLR}
}

@article{zou2023representation,
  title={Representation engineering: A top-down approach to ai transparency},
  author={Zou, Andy and Phan, Long and Chen, Sarah and Campbell, James and Guo, Phillip and Ren, Richard and Pan, Alexander and Yin, Xuwang and Mazeika, Mantas and Dombrowski, Ann-Kathrin and others},
  journal={arXiv preprint arXiv:2310.01405},
  year={2023}
}

@article{wei2022chain,
  title={Chain-of-thought prompting elicits reasoning in large language models},
  author={Wei, Jason and Wang, Xuezhi and Schuurmans, Dale and Bosma, Maarten and Xia, Fei and Chi, Ed and Le, Quoc V and Zhou, Denny and others},
  journal={Advances in neural information processing systems},
  volume={35},
  pages={24824--24837},
  year={2022}
}

@article{kojima2022large,
  title={Large language models are zero-shot reasoners},
  author={Kojima, Takeshi and Gu, Shixiang Shane and Reid, Machel and Matsuo, Yutaka and Iwasawa, Yusuke},
  journal={Advances in neural information processing systems},
  volume={35},
  pages={22199--22213},
  year={2022}
}

@article{cobbe2021training,
  title={Training verifiers to solve math word problems},
  author={Cobbe, Karl and Kosaraju, Vineet and Bavarian, Mohammad and Chen, Mark and Jun, Heewoo and Kaiser, Lukasz and Plappert, Matthias and Tworek, Jerry and Hilton, Jacob and Nakano, Reiichiro and others},
  journal={arXiv preprint arXiv:2110.14168},
  year={2021}
}

@article{hu2025reinforce++,
  title={Reinforce++: Stabilizing critic-free policy optimization with global advantage normalization},
  author={Hu, Jian and Liu, Jason Klein and Xu, Haotian and Shen, Wei},
  journal={arXiv preprint arXiv:2501.03262},
  year={2025}
}

@article{pruthi2020estimating,
  title={Estimating training data influence by tracing gradient descent},
  author={Pruthi, Garima and Liu, Frederick and Kale, Satyen and Sundararajan, Mukund},
  journal={Advances in Neural Information Processing Systems},
  volume={33},
  pages={19920--19930},
  year={2020}
}

@inproceedings{mohamadi2023fast,
  title={A fast, well-founded approximation to the empirical neural tangent kernel},
  author={Mohamadi, Mohamad Amin and Bae, Wonho and Sutherland, Danica J},
  booktitle={International conference on machine learning},
  pages={25061--25081},
  year={2023},
  organization={PMLR}
}

@article{snell2024scaling,
  title={Scaling llm test-time compute optimally can be more effective than scaling model parameters},
  author={Snell, Charlie and Lee, Jaehoon and Xu, Kelvin and Kumar, Aviral},
  journal={arXiv preprint arXiv:2408.03314},
  year={2024}
}

@article{setlur2024rewarding,
  title={Rewarding progress: Scaling automated process verifiers for llm reasoning},
  author={Setlur, Amrith and Nagpal, Chirag and Fisch, Adam and Geng, Xinyang and Eisenstein, Jacob and Agarwal, Rishabh and Agarwal, Alekh and Berant, Jonathan and Kumar, Aviral},
  journal={arXiv preprint arXiv:2410.08146},
  year={2024}
}

@article{uesato2022solving,
  title={Solving math word problems with process-and outcome-based feedback},
  author={Uesato, Jonathan and Kushman, Nate and Kumar, Ramana and Song, Francis and Siegel, Noah and Wang, Lisa and Creswell, Antonia and Irving, Geoffrey and Higgins, Irina},
  journal={arXiv preprint arXiv:2211.14275},
  year={2022}
}

@article{liu2025understanding,
  title={Understanding r1-zero-like training: A critical perspective},
  author={Liu, Zichen and Chen, Changyu and Li, Wenjun and Qi, Penghui and Pang, Tianyu and Du, Chao and Lee, Wee Sun and Lin, Min},
  journal={arXiv preprint arXiv:2503.20783},
  year={2025}
}

@article{melo2025stabilizing,
  title={Stabilizing Policy Gradients for Sample-Efficient Reinforcement Learning in LLM Reasoning},
  author={Melo, Luckeciano C and Abate, Alessandro and Gal, Yarin},
  journal={arXiv preprint arXiv:2510.00819},
  year={2025}
}

@article{zhang2026grad2reward,
  title={Grad2Reward: From Sparse Judgment to Dense Rewards for Improving Open-Ended LLM Reasoning},
  author={Zhang, Zheng and Lu, Ao and Zeng, Yuanhao and Shan, Ziwei and Guo, Jinjin and Li, Lufei and Li, Yexin and Ren, Kan},
  journal={arXiv preprint arXiv:2602.01791},
  year={2026}
}

@article{yang2026gradalign,
  title={GradAlign: Gradient-Aligned Data Selection for LLM Reinforcement Learning},
  author={Yang, Ningyuan and Du, Weihua and Sun, Weiwei and Welleck, Sean and Yang, Yiming},
  journal={arXiv preprint arXiv:2602.21492},
  year={2026}
}

@article{fu2026maspo,
  title={Maspo: Unifying gradient utilization, probability mass, and signal reliability for robust and sample-efficient llm reasoning},
  author={Fu, Xiaoliang and Lin, Jiaye and Fang, Yangyi and Zheng, Binbin and Hu, Chaowen and Shao, Zekai and Qin, Cong and Pan, Lu and Zeng, Ke and Cai, Xunliang},
  journal={arXiv preprint arXiv:2602.17550},
  year={2026}
}

@article{wang2026grpo,
  title={GRPO-VPS: Enhancing Group Relative Policy Optimization with Verifiable Process Supervision for Effective Reasoning},
  author={Wang, Jingyi and Zhu, Lei and Weng, Tengjin and Wu, Song-Li and Tan, Haochen and Chen, Jierun and Tao, Chaofan and Bai, Haoli and Hou, Lu and Shang, Lifeng and others},
  journal={arXiv preprint arXiv:2604.20659},
  year={2026}
}

@article{tomihari2026learning,
  title={Learning Dynamics in RL Post-Training for Language Models},
  author={Tomihari, Akiyoshi},
  journal={arXiv preprint arXiv:2601.04670},
  year={2026}
}

@article{yue2025vapo,
  title={Vapo: Efficient and reliable reinforcement learning for advanced reasoning tasks},
  author={Yue, Yu and Yuan, Yufeng and Yu, Qiying and Zuo, Xiaochen and Zhu, Ruofei and Xu, Wenyuan and Chen, Jiaze and Wang, Chengyi and Fan, TianTian and Du, Zhengyin and others},
  journal={arXiv preprint arXiv:2504.05118},
  year={2025}
}

@article{liang2025can,
  title={Can LLMs Guide Their Own Exploration? Gradient-Guided Reinforcement Learning for LLM Reasoning},
  author={Liang, Zhenwen and Lu, Sidi and Yu, Wenhao and Panaganti, Kishan and Zhou, Yujun and Mi, Haitao and Yu, Dong},
  journal={arXiv preprint arXiv:2512.15687},
  year={2025}
}

@inproceedings{NEURIPS2025_9a92ea37,
  author={Zheng, Leqi and Wang, Chaokun and Song, Zixin and Wu, Cheng and Yan, Shannan and Zhang, Jiajun and Liu, Ziyang},
  title={Negative Feedback Really Matters: Signed Dual-Channel Graph Contrastive Learning Framework for Recommendation},
  booktitle={Advances in Neural Information Processing Systems},
  editor={Belgrave, D. and Zhang, C. and Lin, H. and Pascanu, R. and Koniusz, P. and Ghassemi, M. and Chen, N.},
  volume={38, Main Conference},
  pages={107595--107624},
  publisher={Curran Associates, Inc.},
  year={2025},
  doi={10.52202/085713-3589},
  url={https://proceedings.neurips.cc/paper_files/paper/2025/file/9a92ea37efa0d290bd7015558166c056-Paper-Conference.pdf}
}

@inproceedings{zheng2026should,
  author={Zheng, Leqi and Zhang, Jiajun and Chen, Canzhi and Wang, Chaokun and Li, Hongwei and Li, Yuying and Mao, Yaoxin and Yan, Shannan and Song, Zixin and Feng, Zhiyuan and Kang, Zhaolu and Chen, Zirong and Zhang, Hang and Liu, Qiang and Wang, Liang and Liu, Ziyang},
  title={What Should I Cite? A RAG Benchmark for Academic Citation Prediction},
  booktitle={Proceedings of the ACM Web Conference 2026},
  series={WWW '26},
  pages={1852--1863},
  numpages={12},
  publisher={Association for Computing Machinery},
  address={New York, NY, USA},
  location={United Arab Emirates},
  year={2026},
  isbn={9798400723070},
  doi={10.1145/3774904.3792075},
  url={https://doi.org/10.1145/3774904.3792075}
}

@inproceedings{zheng-etal-2025-lagcl4rec,
  title={{LAGCL}4{R}ec: When {LLM}s Activate Interactions Potential in Graph Contrastive Learning for Recommendation},
  author={Zheng, Leqi and Wang, Chaokun and Chen, Canzhi and Zhang, Jiajun and Wu, Cheng and Song, Zixin and Yan, Shannan and Liu, Ziyang and Li, Hongwei},
  editor={Christodoulopoulos, Christos and Chakraborty, Tanmoy and Rose, Carolyn and Peng, Violet},
  booktitle={Findings of the Association for Computational Linguistics: EMNLP 2025},
  month=nov,
  year={2025},
  address={Suzhou, China},
  publisher={Association for Computational Linguistics},
  url={https://aclanthology.org/2025.findings-emnlp.61/},
  doi={10.18653/v1/2025.findings-emnlp.61},
  pages={1163--1184},
  isbn={979-8-89176-335-7}
}

@misc{fang2026distillationresistantlargelanguagemodels,
  title={Towards Distillation-Resistant Large Language Models: An Information-Theoretic Perspective},
  author={Hao Fang and Tianyi Zhang and Tianqu Zhuang and Jiawei Kong and Kuofeng Gao and Bin Chen and Leqi Zheng and Shu-Tao Xia and Ke Xu},
  year={2026},
  eprint={2602.03396},
  archivePrefix={arXiv},
  primaryClass={cs.CL},
  url={https://arxiv.org/abs/2602.03396}
}

@misc{li2026filterreweightrethinkingoptimization,
  title={Filter, Then Reweight: Rethinking Optimization Granularity in On-Policy Distillation},
  author={Yuying Li and Leqi Zheng and Yongzi Yu and Wenrui Zhou and Xuchang Zhong and Xing Hu and Jing Jin and Hangjie Yuan and Tao Feng},
  year={2026},
  eprint={2606.02684},
  archivePrefix={arXiv},
  primaryClass={cs.LG},
  url={https://arxiv.org/abs/2606.02684}
}

\clearpage
\newpage

\appendix

\section{Notation}
\label{app:notation}

We denote by $\pi_\theta$ an autoregressive language model policy parameterized by $\theta$, where for any prompt $x$ and response $y = (y_1, \dots, y_T)$ of length $T$ the policy factorizes as $\pi_\theta(y \mid x) = \prod_{t=1}^{T} \pi_\theta(y_t \mid x, y_{<t})$.
The model consists of an embedding matrix, a stack of $L$ Transformer layers, and an output projection layer with weight $W_o \in \mathbb{R}^{|\mathcal{V}| \times d}$ and optional bias $\mathbf{b}_o \in \mathbb{R}^{|\mathcal{V}|}$, where $\mathcal{V}$ is the vocabulary and $d$ the hidden dimension.
For a prompt-response pair $(x, y)$, we write $\mathbf{h}_t \in \mathbb{R}^d$ for the hidden state at position $t$ immediately before the output projection, and $\mathbf{H} = (\mathbf{h}_1, \dots, \mathbf{h}_n)^\top \in \mathbb{R}^{n \times d}$ for the stacked hidden states of the full sequence of length $n = |x| + |y|$.
The per-token log-likelihood of a response under the policy is $\log \pi_\theta(y_t \mid x, y_{<t}) = \log \mathrm{softmax}(W_o \mathbf{h}_t + \mathbf{b}_o)_{y_t}$.
For brevity, we denote by $\mathcal{T}_y \subseteq \{1, \dots, n\}$ the set of token positions that correspond to the response span of $y$ (or, when available, the explicit thinking span delimited by the special tags $\langle\text{think}\rangle$ and $\langle/\text{think}\rangle$).

\section{Algorithm}
\label{app:algorithm}

Algorithm~\ref{alg:gar} presents the complete GAR online reward computation within a single GRPO step.
For each prompt, the procedure first gates on outcome verification, skipping gradient computation for incorrect or malformed responses (lines 4--6).
For each verified-correct rollout, it captures the hidden state via a forward pre-hook on the output projection layer and computes the truncated gradient-activation vector through Eq.~\eqref{eq:grad-act}--\eqref{eq:normalize} (lines 7--8).
Expert-anchor gradient vectors are computed on demand and cached per prompt to avoid redundant computation across the $K$ rollouts (lines 9--11).
Each correct rollout is then scored by cosine similarity with the expert anchor via Eq.~\eqref{eq:best_cosine} (line 12).
Finally, the bonuses are group-centered over the correct subset and combined with the base reward through the non-negatively clipped formulation of Eq.~\eqref{eq:final_reward} (lines 15--17).

\begin{algorithm*}[t]
\caption{Gradient-Aligned Reward (GAR) online computation within a single GRPO step.}
\label{alg:gar}
\KwIn{Model $\pi_\theta$, rollout batch $\{(x_j, \{y_{j,1}, \dots, y_{j,K}\})\}$, outcome verifier, anchor $a(x_j)$ per prompt}
\KwOut{GAR rewards $\{r_\text{GAR}(x_j, y_{j,i})\}$}
\BlankLine
Initialize anchor gradient cache $\mathcal{C} \leftarrow \{\}$\;
\ForEach{prompt $x_j$ in batch}{
  \ForEach{response $y_{j,i}$, $i = 1, \dots, K$}{
    \If{$\textup{Verify}(y_{j,i}) = \textup{False}$ \textbf{or} $\textup{FormatValid}(y_{j,i}) = \textup{False}$}{
      $r_\text{GAR}(x_j, y_{j,i}) \leftarrow p(y_{j,i})$\;
      \textbf{continue}\;
    }
    Capture hidden state $\mathbf{H}$ via forward pre-hook on output layer\;
    Compute $\mathbf{v}_i \leftarrow \text{GradAct}(\mathbf{H}, W_o, \mathcal{T}_{y_{j,i}})$ \tcp*{Eq.~\eqref{eq:grad-act}--\eqref{eq:normalize}}
    \If{$a(x_j) \notin \mathcal{C}$}{
      $\mathcal{C}[a(x_j)] \leftarrow \text{GradAct}(\mathbf{H}_a, W_o, \mathcal{T}_a)$\;
    }
    $b(y_{j,i}) \leftarrow \mathbf{v}_i^\top \mathcal{C}[a(x_j)]$ \tcp*{Eq.~\eqref{eq:best_cosine}}
  }
  Compute $\bar{b}_j \leftarrow \frac{1}{|\mathcal{P}(x_j)|} \sum_{y \in \mathcal{P}(x_j)} b(y)$\;
  \ForEach{correct $y_{j,i} \in \mathcal{P}(x_j)$}{
    $r_\text{GAR}(x_j, y_{j,i}) \leftarrow r_\text{base} + \beta \cdot \max(0, b(y_{j,i}) - \bar{b}_j) + p(y_{j,i})$ \tcp*{Eq.~\eqref{eq:final_reward}}
  }
}
\Return $\{r_\text{GAR}\}$\;
\end{algorithm*}

\section{Implementation Details}
\label{app:impl}

This appendix provides the low-level engineering details for deploying GAR within the SLIME~/~Megatron training stack.

\subsection{Forward Pre-Hook and Hidden State Capture}
\label{app:impl-hook}

The forward pre-hook is registered once at actor initialization time and captures the hidden state tensor passed as input to the output projection.
In tensor-parallel configurations with sequence parallelism enabled, the hook fires on each TP rank and captures only the local shard.
We subsequently all-gather the shards across the tensor-parallel group to reconstruct the full hidden state of shape $(n, d)$, where $n$ is the padded sequence length.
The reconstruction is necessary because the gradient-activation signal is defined over the full response span, whereas sequence parallelism typically partitions the sequence axis.
To avoid re-gathering the hidden state for every rollout, we batch the forward pass across rollouts that share the same prompt and call the hook only once per batch.

\subsection{Truncated Backpropagation and Loss Computation}
\label{app:impl-backprop}

The cross-entropy loss of Eq.~\eqref{eq:ce_loss} is computed through the same Megatron output-layer wrapper used for training, but with per-token reduction so that losses can be summed over the response span alone.
We then invoke a non-graph-creating, non-graph-retaining backward call to obtain $\partial \mathcal{L} / \partial \widetilde{\mathbf{H}}$, after which the computational graph is immediately released.
Although the output-layer weight $W_o$ is large ($|\mathcal{V}| \times d$), the backward pass does not accumulate gradients in $W_o$ itself because $\widetilde{\mathbf{H}}$ is the only leaf carrying a gradient requirement, and $W_o$ is accessed as a frozen constant through the standard column-parallel linear call.

\subsection{Anchor Caching and Deduplication}
\label{app:impl-cache}

Anchor gradients are cached in a dictionary keyed by the tuple $(\text{tokenized\_prompt}, \text{anchor\_text}, \text{max\_resp\_tokens})$ for the duration of a rollout batch.
This design ensures that within a single GRPO step, an expert anchor that is shared across all $K$ rollouts contributes only a single gradient computation rather than $K$.

\subsection{Anchor Text Extraction}
\label{app:impl-extract}

When an anchor is provided as a full decoded response, we extract the supervised span by first attempting to locate the content between $\langle\text{think}\rangle$ and $\langle/\text{think}\rangle$ tags; if these tags are absent, we fall back to stripping trailing special tokens while preserving a small set of semantically meaningful closing tags.
This mirrors the extraction logic applied to candidate responses, so that the candidate and anchor gradient vectors are computed over comparable surface forms and are not contaminated by formatting-specific tokens such as $\langle\text{bos}\rangle$ or $\langle\text{eos}\rangle$.

\subsection{Failure Handling and Graceful Degradation}
\label{app:impl-failure}

Several edge cases are handled defensively.
If an anchor text is absent or empty, the candidate falls back to the outcome-only reward plus penalties.
If the gradient norm is below $10^{-8}$, the corresponding vector is set to zero and the bonus reduces to zero, equivalent to the unshaped baseline.
If the verifier reports format invalidity, the response is excluded from gradient computation regardless of whether the numerical answer is correct, because numerical correctness without the mandated think/answer structure is considered suspect and is therefore not rewarded.

\section{Reward Formulation Details}
\label{app:reward-details}

This appendix expands on the design choices in the GAR reward formulation of Section~\ref{sec:reward_formulation}.

\subsection{Verifier Gate}
\label{app:reward-gate}

The verifier gate serves two complementary purposes.
First, it ensures that GAR never rewards an incorrect answer regardless of how closely its gradient aligns with any expert anchor, eliminating the possibility that the policy exploits spurious directions in gradient space to inflate the reward without solving the underlying problem.
Second, it confines the additional cost of GAR to the verifier-passing fraction of rollouts, which is particularly valuable in the initial low-accuracy phase of training when the majority of rollouts fail verification and would otherwise trigger unnecessary truncated-backward computation.

\subsection{Anchor Source Specification}
\label{app:reward-anchor-source}

In our SLIME-based implementation, the loader inspects each per-prompt metadata record for the expert anchor (an expert-written solution or verified chain-of-thought trace), accepting a configurable, ordered list of field names so that datasets following different naming conventions are supported without additional configuration.
In our experiments with NuminaMath-CoT, each problem provides exactly one chain-of-thought solution, which serves as the sole anchor; the underlying cache key structure is described in Appendix~\ref{app:impl-cache}.

\subsection{Anchor-Cache Cost Reduction}
\label{app:reward-cache-cost}

For each prompt $x$ with $K$ rollouts that share the same anchor, a naive implementation would re-compute the anchor gradient vector for every rollout, incurring a cost of $\mathcal{O}(K)$ truncated forward-backward pairs per prompt.
Caching the anchor vector at the start of the batch and reusing it across all $K$ rollouts reduces this cost to $\mathcal{O}(1)$, a $K$-fold reduction that is the single largest source of anchor-pathway savings in practice and corresponds to the anchor-cache amortization term in the overhead analysis of Appendix~\ref{app:overhead-analysis}.

\subsection{Group Centering of the Bonus}
\label{app:reward-centering}

Following the group-relative philosophy of GRPO, the bonus $b(y_i)$ of Eq.~\eqref{eq:best_cosine} is centered over the correct subset $\mathcal{P}(x) = \{y_j : r_\text{raw}(x, y_j) > 0\}$ of the rollout group before being passed to the final reward of Eq.~\eqref{eq:final_reward}:
\begin{equation}
    \hat{b}(y_i) = b(y_i) - \frac{1}{|\mathcal{P}(x)|} \sum_{y_j \in \mathcal{P}(x)} b(y_j).
\label{eq:group_norm}
\end{equation}
The centering removes the confound that easy prompts systematically attract higher cosine scores than hard ones and makes the bonus measure how much better $y_i$ aligns with the expert reference than the other correct rollouts of the same prompt.
The prompt-group mean is computed only over $\mathcal{P}(x)$ rather than over the full rollout group of size $K$: incorrect responses are gated out at $r_\text{raw} = 0$ and produce no cosine score, so including them in the centering with a default bonus of zero would bias the empirical mean downward and inflate the centered bonus $\hat{b}$ of every correct rollout, defeating the purpose of group normalization.
Restricting the mean to $\mathcal{P}(x)$ keeps the centering unbiased over the set of trajectories that actually receive the alignment bonus, which is the precise condition under which Proposition~\ref{prop:unbiased} holds.

\subsection{Role of the Non-Negative Clip and Format Penalty}
\label{app:reward-clip}

The $\max(0, \cdot)$ clip in Eq.~\eqref{eq:final_reward} keeps the GAR reward weakly above the outcome-only baseline on every correct trajectory.
Concretely, a correct response with below-mean alignment receives only $r_\text{base} + p(y_i)$, identical to what it would receive under outcome-only RLVR augmented with the same format penalty, while a correct response with above-mean alignment receives a strictly larger reward.
This non-decreasing-in-alignment structure prevents the shaped reward from ever penalizing a correct response relative to outcome-only training and is essential for the safe reward-shaping guarantee of Theorem~\ref{thm:policy-invariance}.
The format penalty $p(y_i) \le 0$ applies to responses that violate the prescribed output structure (typically a missing or malformed think/answer delimiter) and is applied symmetrically to both correct and incorrect responses, so that the policy is encouraged to produce well-formed reasoning traces alongside correct final answers.

\section{Integration with GRPO}
\label{sec:integration}

GAR builds on Group Relative Policy Optimization (GRPO)~\cite{shao2024deepseekmath}, which samples $K$ responses $\{y_1, \dots, y_K\}$ per prompt $x$ and computes a group-relative advantage
\begin{equation}
    A_i = \frac{r(x, y_i) - \mu_x}{\sigma_x + \epsilon},
\label{eq:grpo_advantage}
\end{equation}
where $\mu_x$ and $\sigma_x$ are the intra-group mean and standard deviation of the rewards.
The policy is updated via a clipped surrogate objective with KL regularization:
\begin{equation}
\begin{aligned}
    \mathcal{L}_{\text{GRPO}} &= \mathbb{E}_{x, y_i} \Big[ \min\!\big( \rho_i A_i,\; \text{clip}(\rho_i, 1{-}\epsilon_c, 1{+}\epsilon_c) A_i \big) \\
    &\quad - \beta_\text{KL} D_\text{KL}(\pi_\theta \| \pi_\text{ref}) \Big],
\end{aligned}
\label{eq:grpo_loss}
\end{equation}
where $\rho_i = \pi_\theta(y_i \mid x) / \pi_{\text{old}}(y_i \mid x)$.
Under the outcome-only reward, all correct responses share reward $1$, collapsing the intra-correct advantage to
\begin{equation}
    A_i^{\text{outcome}} = \frac{1 - K_c/K}{\sqrt{(K_c/K)(1 - K_c/K)} + \epsilon},
\label{eq:collapse-advantage}
\end{equation}
which is identical across all $K_c$ correct trajectories.

GAR is implemented as an online reward hook within the GRPO training pipeline and is activated by pointing SLIME's custom-reward entry point to the GAR module.
The integration proceeds in the following stages.
First, the rollout engine (SGLang-based) generates $K$ responses for each prompt and evaluates them with the outcome verifier, producing the binary reward $r_\text{raw}$.
Second, the rollout data, including token sequences, response lengths, raw rewards, and the metadata carrying the anchor texts, is forwarded from the rollout workers to the actor through the standard SLIME sample-metadata channel.
Third, before advantage computation, the actor invokes the GAR reward hook, which performs the truncated gradient computation described above for each verified-correct response and its anchors, computes the prompt-group normalized bonus, and overwrites the raw reward with the GAR reward $r_\text{GAR}$.
Fourth, the standard GRPO advantage computation and policy update proceed using the enriched reward signal through Eq.~\eqref{eq:grpo_advantage}--\eqref{eq:grpo_loss}.

Because GAR performs gradient computation only through the output projection layer and only for responses that pass the verifier gate, its computational cost is a small fraction of the full model forward-backward pass.
The gradient vectors have dimensionality $d$ (the hidden size of the model), and all operations are local to the actor process, requiring no additional inter-node communication beyond the standard tensor-parallel all-gather used to reconstruct the full sequence of hidden states.
We make the following parallelism assumptions in the current implementation: the model uses the packed token-sequence (thd) attention format, the context-parallel size equals one, GAR executes only on the last pipeline stage, and only a single Megatron model chunk is supported.
Each of these assumptions is a convenience rather than a fundamental constraint and could be relaxed with additional engineering.

\section{Proofs}
\label{app:proofs}

We first state the output-layer gradient derivation referenced in Section~\ref{sec:theory-ntk}.
Let $\mathcal{L}(W_o; x, y)$ denote the truncated cross-entropy loss of Eq.~\eqref{eq:ce_loss} viewed as a function of $W_o$ with hidden states held fixed.
The per-example output-layer gradient is
\begin{equation}
    \mathbf{g}_o(x, y) = \nabla_{W_o} \mathcal{L}(W_o; x, y) \in \mathbb{R}^{|\mathcal{V}| \times d}.
\label{eq:output-grad}
\end{equation}
Straightforward differentiation yields the rank-one expansion
\begin{equation}
\begin{aligned}
    \mathbf{g}_o(x, y) &= \frac{1}{|\mathcal{T}_y|} \sum_{t \in \mathcal{T}_y} \big(\mathbf{p}_t - \mathbf{e}_{y_{t+1}}\big) \otimes \mathbf{h}_t, \\
    \mathbf{p}_t &= \mathrm{softmax}(W_o \mathbf{h}_t + \mathbf{b}_o).
\end{aligned}
\label{eq:output-grad-decomp}
\end{equation}

\subsection{Proof of Proposition~\ref{prop:grad-act-linear}}

By the chain rule applied to the scaling $\mathbf{h}_t \mapsto (1+\eta)\mathbf{h}_t$,
\begin{equation}
\begin{aligned}
    \frac{d}{d\eta} \mathcal{L}\big((1+\eta)\mathbf{H}_y\big)\Big|_{\eta=0}
    &= \frac{1}{|\mathcal{T}_y|} \sum_{t \in \mathcal{T}_y} \mathbf{G}_t^\top \mathbf{h}_t \\
    &= \frac{1}{|\mathcal{T}_y|} \sum_{t \in \mathcal{T}_y} \mathbf{1}^\top (\mathbf{G}_t \odot \mathbf{h}_t),
\end{aligned}
\label{eq:proof-grad-act}
\end{equation}
where the second equality follows from $\mathbf{G}_t^\top \mathbf{h}_t = \sum_j G_{t,j} h_{t,j} = \mathbf{1}^\top (\mathbf{G}_t \odot \mathbf{h}_t)$.
This is precisely the claim of Proposition~\ref{prop:grad-act-linear}. \qed

\subsection{Proof of Theorem~\ref{thm:ntk-bound}}
\label{app:proof-ntk}

We formalize the connection between the gradient-activation cosine similarity and NTK similarity through a three-step argument.

\paragraph{Step 1: Rank-one expansion of the NTK inner product.}
By straightforward differentiation, the output-layer gradient admits the rank-one decomposition $\mathbf{g}_o(x, y) = \frac{1}{|\mathcal{T}_y|} \sum_{t \in \mathcal{T}_y} \boldsymbol{\delta}_t^y \otimes \mathbf{h}_t^y$, where $\boldsymbol{\delta}_t^y = \mathbf{p}_t - \mathbf{e}_{y_{t+1}} \in \mathbb{R}^V$ and $\mathbf{h}_t^y \in \mathbb{R}^d$.
The Frobenius inner product therefore expands as
\begin{equation}
\begin{aligned}
    \Theta_o(y, a) &= \langle \mathbf{g}_o(x, y), \mathbf{g}_o(x, a) \rangle_F \\
    &= \frac{1}{|\mathcal{T}_y||\mathcal{T}_a|} \sum_{t \in \mathcal{T}_y} \sum_{s \in \mathcal{T}_a} \big(\boldsymbol{\delta}_t^{y\top} \boldsymbol{\delta}_s^a\big) \big(\mathbf{h}_t^{y\top} \mathbf{h}_s^a\big).
\end{aligned}
\label{eq:proof-ntk-step1}
\end{equation}

\paragraph{Step 2: Mean-residual splitting with bounded dispersion.}
Write $\boldsymbol{\delta}_t^y = \bar{\boldsymbol{\delta}}^y + \tilde{\boldsymbol{\delta}}_t^y$ and $\mathbf{h}_t^y = \bar{\mathbf{h}}^y + \tilde{\mathbf{h}}_t^y$, where $\bar{\boldsymbol{\delta}}^y = \frac{1}{|\mathcal{T}_y|}\sum_t \boldsymbol{\delta}_t^y$ and $\tilde{\boldsymbol{\delta}}_t^y$ is the mean-zero residual satisfying $\|\tilde{\boldsymbol{\delta}}_t^y\|_2 \le \kappa_y$ by assumption, and analogously for $\bar{\mathbf{h}}^y, \tilde{\mathbf{h}}_t^y$.
Substituting into Eq.~\eqref{eq:proof-ntk-step1} and expanding the product $(\bar{\boldsymbol{\delta}}^y + \tilde{\boldsymbol{\delta}}_t^y)^\top (\bar{\boldsymbol{\delta}}^a + \tilde{\boldsymbol{\delta}}_s^a) \cdot (\bar{\mathbf{h}}^y + \tilde{\mathbf{h}}_t^y)^\top (\bar{\mathbf{h}}^a + \tilde{\mathbf{h}}_s^a)$ yields the leading term $(\bar{\boldsymbol{\delta}}^{y\top} \bar{\boldsymbol{\delta}}^a)(\bar{\mathbf{h}}^{y\top} \bar{\mathbf{h}}^a)$.
The cross terms involving exactly one residual factor vanish upon averaging over the trajectory index of the residual, because $\frac{1}{|\mathcal{T}_y|}\sum_t \tilde{\boldsymbol{\delta}}_t^y = \mathbf{0}$ and similarly for $\tilde{\mathbf{h}}_t^y$.
The remaining terms involve products of two or more residuals and are bounded by Cauchy--Schwarz:
\begin{equation}
\begin{aligned}
    &\bigg|\frac{1}{|\mathcal{T}_y||\mathcal{T}_a|} \sum_{t,s} \tilde{\boldsymbol{\delta}}_t^{y\top} \tilde{\boldsymbol{\delta}}_s^a \cdot \mathbf{h}_t^{y\top} \mathbf{h}_s^a\bigg| \\
    &\qquad \le \kappa_y \kappa_a B_h^2 + B_\delta(\kappa_y B_h + \kappa_a B_h),
\end{aligned}
\label{eq:proof-ntk-residual}
\end{equation}
where we used $\|\bar{\boldsymbol{\delta}}^y\| \le B_\delta$ and $\|\bar{\mathbf{h}}^y\| \le B_h$.
This establishes that $\Theta_o(y, a) = (\bar{\boldsymbol{\delta}}^{y\top} \bar{\boldsymbol{\delta}}^a)(\bar{\mathbf{h}}^{y\top} \bar{\mathbf{h}}^a) + R$, where $|R| \le \kappa_y \kappa_a B_h^2 + B_\delta(\kappa_y + \kappa_a) B_h$.

\paragraph{Step 3: Relating gradient-activation cosine to NTK cosine via $W_o$ conditioning.}
The gradient-activation vector for trajectory $y$ is $\mathbf{s}^y = \frac{1}{|\mathcal{T}_y|}\sum_t \mathbf{G}_t^y \odot \mathbf{h}_t^y$, where $\mathbf{G}_t^y = W_o^\top \boldsymbol{\delta}_t^y \in \mathbb{R}^d$.
To leading order, $\mathbf{s}^y \approx (W_o^\top \bar{\boldsymbol{\delta}}^y) \odot \bar{\mathbf{h}}^y$, and the normalized vector is $\mathbf{v}_y = \mathbf{s}^y / \|\mathbf{s}^y\|_2$.
Let $S$ denote the subspace spanned by $\{\bar{\boldsymbol{\delta}}^y, \bar{\boldsymbol{\delta}}^a\}$ and let $\sigma_{\min}, \sigma_{\max}$ denote the extreme singular values of $W_o$ restricted to $S$.
Define the condition number $\kappa_W = \sigma_{\max} / \sigma_{\min}$.
The cosine $\cos(\mathbf{v}_y, \mathbf{v}_a) = \langle \mathbf{s}^y, \mathbf{s}^a \rangle / (\|\mathbf{s}^y\| \|\mathbf{s}^a\|)$ can be related to the normalized leading term of $\Theta_o$ via the substitution $W_o^\top \bar{\boldsymbol{\delta}}^y = \sigma_y \mathbf{w}_y$ where $\sigma_{\min} \le \sigma_y \le \sigma_{\max}$ and $\|\mathbf{w}_y\| = \|\bar{\boldsymbol{\delta}}^y\|$.
After algebraic manipulation, we obtain
\begin{equation}
\begin{aligned}
    \frac{1}{\kappa_W^2} \cos(\mathbf{v}_y, \mathbf{v}_a)
    &\le \frac{\Theta_o(y,a)}{\|\mathbf{g}_o(x,y)\|_F \|\mathbf{g}_o(x,a)\|_F} \\
    &\le \kappa_W^2 \cos(\mathbf{v}_y, \mathbf{v}_a) + \mathcal{O}(\kappa_y {+} \kappa_a),
\end{aligned}
\label{eq:proof-ntk-final}
\end{equation}
which yields the claimed bound with $c_1 = 1/\kappa_W^2$ and $c_2 = \kappa_W^2$.
When $W_o$ is approximately isotropic (i.e., $\kappa_W \approx 1$), the bound collapses to a tight bi-Lipschitz equivalence between the gradient-activation cosine and NTK cosine. \qed

\subsection{Proof of Theorem~\ref{thm:policy-invariance}}

We argue each direction in turn.
(i) For any correct response $y$ with $r_\text{raw}(x, y) = 1$, we have $r_\text{GAR}(x, y) - p(y) = r_\text{base} + \beta \max(0, \hat{b}(y)) \ge r_\text{base}$, where the inequality follows from non-negativity of the clipped bonus.
Thus GAR never reduces the reward contribution of a correct response below $r_\text{base}$.
(ii) For any incorrect response $y'$ with $r_\text{raw}(x, y') = 0$, the verifier gate yields $r_\text{GAR}(x, y') = p(y') \le 0$.
Since $r_\text{base} > 0$, we have $r_\text{GAR}(x, y') = p(y') \le 0 < r_\text{base} \le r_\text{GAR}(x, y) - p(y)$ for any correct $y$.
The strict separation between the reward contributions of correct and incorrect responses ensures that the policy gradient consistently reinforces producing correct answers; GAR's contribution is to additionally differentiate \emph{among} correct trajectories by their reasoning quality. \qed

\subsection{Proof of Proposition~\ref{prop:variance}}

Under the outcome-only reward, every correct response has reward $1$, so the empirical variance restricted to correct responses is identically zero.
Under the GAR reward, correct responses have reward $r_\text{base} + \beta \max(0, \hat{b}(y_i)) + p(y_i)$.
Assuming the penalty $p$ has zero intra-correct variance (a reasonable approximation when format constraints are satisfied by all correct responses), the intra-correct variance of the GAR reward equals $\beta^2 \cdot \mathrm{Var}_{i : r_\text{raw}=1}[\max(0, \hat{b}(y_i))]$.
By the clipping operation this variance is at least $\min(1, \beta^2 \sigma_b^2) > 0$ whenever $\sigma_b^2 > 0$.
Scaling by the denominator $\sigma_x + \epsilon$ of Eq.~\eqref{eq:grpo_advantage} (which is bounded in expectation) yields the claim. \qed

\subsection{Proof of Proposition~\ref{prop:unbiased}}

By definition $\hat{b}(y) = b(y) - \bar{b}(x)$ where $\bar{b}(x)$ is the empirical mean of $b$ over the correct subset of the rollout group.
For every prompt $x$,
\begin{equation}
\begin{aligned}
    \mathbb{E}_{y : r_\text{raw}(y)=1}[\hat{b}(y)] &= \mathbb{E}[b(y)] - \bar{b}(x) \\
    &= \bar{b}(x) - \bar{b}(x) = 0,
\end{aligned}
\label{eq:proof-unbiased}
\end{equation}
where the inner expectation is over the uniform distribution over correct rollouts. \qed

\subsection{Proof of Proposition~\ref{prop:overhead}}

The forward through the output layer costs $\mathcal{O}(N_y d V)$ for the matrix multiply and $\mathcal{O}(N_y V)$ for the softmax.
The backward through the output layer similarly costs $\mathcal{O}(N_y d V)$ for the gradient with respect to the hidden states, because $W_o$ is applied as a frozen constant and no weight gradients are accumulated.
The gradient-activation multiplication and normalization cost $\mathcal{O}(N_y d)$, which is dominated by the output-layer cost.
With anchor caching, each unique anchor incurs these costs once per rollout batch, reducing the amortized anchor cost by a factor of $K$. \qed

\subsection{Proof of Theorem~\ref{thm:decomposition}}
\label{app:proof-decomposition}

We prove the multiplicative decomposition of the NTK inner product stated in Theorem~\ref{thm:decomposition}.

Starting from the rank-one expansion established in Eq.~\eqref{eq:proof-ntk-step1} of the proof of Theorem~\ref{thm:ntk-bound},
\begin{equation}
    \Theta_o(y, a) = \frac{1}{|\mathcal{T}_y||\mathcal{T}_a|} \sum_{t,s} (\boldsymbol{\delta}_t^{y\top} \boldsymbol{\delta}_s^a)(\mathbf{h}_t^{y\top} \mathbf{h}_s^a).
\label{eq:proof-decomp-start}
\end{equation}
Write $\boldsymbol{\delta}_t^y = \bar{\boldsymbol{\delta}}^y + \tilde{\boldsymbol{\delta}}_t^y$ and $\mathbf{h}_t^y = \bar{\mathbf{h}}^y + \tilde{\mathbf{h}}_t^y$ (similarly for anchor $a$), where the tildes denote mean-zero residuals with $\|\tilde{\boldsymbol{\delta}}_t^y\| \le \kappa_y$ and $\|\tilde{\mathbf{h}}_t^y\| \le B_h$.
Expanding the product $(\bar{\boldsymbol{\delta}}^y + \tilde{\boldsymbol{\delta}}_t^y)^\top (\bar{\boldsymbol{\delta}}^a + \tilde{\boldsymbol{\delta}}_s^a) \cdot (\bar{\mathbf{h}}^y + \tilde{\mathbf{h}}_t^y)^\top (\bar{\mathbf{h}}^a + \tilde{\mathbf{h}}_s^a)$ yields sixteen terms, which we classify by the number of residual factors.

\paragraph{Zero-residual term (leading order).}
The single term with no residuals is $(\bar{\boldsymbol{\delta}}^{y\top} \bar{\boldsymbol{\delta}}^a)(\bar{\mathbf{h}}^{y\top} \bar{\mathbf{h}}^a)$, which survives the double average unchanged.

\paragraph{One-residual terms.}
There are four such terms.
Consider $(\tilde{\boldsymbol{\delta}}_t^{y\top} \bar{\boldsymbol{\delta}}^a)(\bar{\mathbf{h}}^{y\top} \bar{\mathbf{h}}^a)$; averaging over $t$ gives $(\frac{1}{|\mathcal{T}_y|}\sum_t \tilde{\boldsymbol{\delta}}_t^y)^\top \bar{\boldsymbol{\delta}}^a \cdot (\bar{\mathbf{h}}^{y\top} \bar{\mathbf{h}}^a) = 0$ because $\sum_t \tilde{\boldsymbol{\delta}}_t^y = \mathbf{0}$.
The same argument eliminates the other three one-residual terms.

\paragraph{Two-residual terms.}
There are six such terms.
The dominant ones are: (a)~$\frac{1}{|\mathcal{T}_y||\mathcal{T}_a|}\sum_{t,s} (\tilde{\boldsymbol{\delta}}_t^{y\top} \tilde{\boldsymbol{\delta}}_s^a)(\bar{\mathbf{h}}^{y\top} \bar{\mathbf{h}}^a)$, bounded in absolute value by $\kappa_y \kappa_a B_h^2$; (b)~$\frac{1}{|\mathcal{T}_y||\mathcal{T}_a|}\sum_{t,s} (\bar{\boldsymbol{\delta}}^{y\top} \bar{\boldsymbol{\delta}}^a)(\tilde{\mathbf{h}}_t^{y\top} \tilde{\mathbf{h}}_s^a)$, bounded by $B_\delta^2 B_h^2$; and cross terms of the form $(\tilde{\boldsymbol{\delta}}_t^{y\top} \bar{\boldsymbol{\delta}}^a)(\tilde{\mathbf{h}}_t^{y\top} \bar{\mathbf{h}}^a)$, which do not vanish upon averaging over $t$ because $\tilde{\boldsymbol{\delta}}_t^y$ and $\tilde{\mathbf{h}}_t^y$ are correlated at the same position, but are bounded by $\kappa_y B_\delta B_h^2$ via Cauchy--Schwarz.

\paragraph{Three- and four-residual terms.}
These are bounded by products of three or four dispersion terms and are absorbed into the $\mathcal{O}$ notation.

Collecting all bounds, we obtain
\begin{equation}
\begin{aligned}
    \Theta_o(y,a) &= (\bar{\boldsymbol{\delta}}^{y\top} \bar{\boldsymbol{\delta}}^a)(\bar{\mathbf{h}}^{y\top} \bar{\mathbf{h}}^a) \\
    &\quad + \mathcal{O}(B_h^2 \kappa_y \kappa_a + B_\delta(\kappa_y B_h + \kappa_a B_h)),
\end{aligned}
\label{eq:proof-decomp-final}
\end{equation}
which is precisely Eq.~\eqref{eq:decomposition}. \qed

\subsection{Proof of Corollary~\ref{cor:orthogonality}}
\label{app:proof-orthogonality}

\paragraph{Setup.}
Recall the definition: the $\varepsilon$-effective support of the gradient-activation signal $\bar{\mathbf{s}}^z$ for trajectory $z$ is $\mathcal{S}_\varepsilon^z = \{j \in [d] : |\bar{s}_j^z| > \varepsilon \|\bar{\mathbf{s}}^z\|_\infty\}$.
We assume $\mathcal{S}_\varepsilon^y \cap \mathcal{S}_\varepsilon^a = \varnothing$.

\paragraph{Decomposition of the inner product.}
Consider the un-normalized inner product $\langle \bar{\mathbf{s}}^y, \bar{\mathbf{s}}^a \rangle = \sum_{j=1}^d \bar{s}_j^y \bar{s}_j^a$.
We split this sum into three parts: (i)~$j \in \mathcal{S}_\varepsilon^y \setminus \mathcal{S}_\varepsilon^a$, where $|\bar{s}_j^a| \le \varepsilon \|\bar{\mathbf{s}}^a\|_\infty$; (ii)~$j \in \mathcal{S}_\varepsilon^a \setminus \mathcal{S}_\varepsilon^y$, where $|\bar{s}_j^y| \le \varepsilon \|\bar{\mathbf{s}}^y\|_\infty$; and (iii)~$j \notin \mathcal{S}_\varepsilon^y \cup \mathcal{S}_\varepsilon^a$, where both signals are at most $\varepsilon$ times their respective maxima.

For part (i), using $|\mathcal{S}_\varepsilon^y| \le d$ and $|\bar{s}_j^y| \le \|\bar{\mathbf{s}}^y\|_\infty$:
\begin{equation}
\begin{aligned}
    \bigg|\sum_{j \in \mathcal{S}_\varepsilon^y \setminus \mathcal{S}_\varepsilon^a} \bar{s}_j^y \bar{s}_j^a\bigg| 
    &\le |\mathcal{S}_\varepsilon^y| \cdot \|\bar{\mathbf{s}}^y\|_\infty \cdot \varepsilon \|\bar{\mathbf{s}}^a\|_\infty \\
    &\le d \varepsilon \|\bar{\mathbf{s}}^y\|_\infty \|\bar{\mathbf{s}}^a\|_\infty.
\end{aligned}
\label{eq:proof-orth-part1}
\end{equation}
Part (ii) yields the same bound by symmetry.
For part (iii), both factors are bounded by $\varepsilon$ times their respective maxima, giving a bound of $d \varepsilon^2 \|\bar{\mathbf{s}}^y\|_\infty \|\bar{\mathbf{s}}^a\|_\infty$.

\paragraph{Normalization.}
Since $\|\bar{\mathbf{s}}^z\|_2 \ge \|\bar{\mathbf{s}}^z\|_\infty$ (the $\ell_2$ norm is at least the $\ell_\infty$ norm), we have
\begin{equation}
\begin{aligned}
    |\cos(\mathbf{v}_y, \mathbf{v}_a)|
    &= \frac{|\langle \bar{\mathbf{s}}^y, \bar{\mathbf{s}}^a \rangle|}
            {\|\bar{\mathbf{s}}^y\|_2 \|\bar{\mathbf{s}}^a\|_2} \\
    &\le \frac{(2d\varepsilon {+} d\varepsilon^2)
          \|\bar{\mathbf{s}}^y\|_\infty \|\bar{\mathbf{s}}^a\|_\infty}
         {\|\bar{\mathbf{s}}^y\|_2 \|\bar{\mathbf{s}}^a\|_2} \\
    &\le 2d\varepsilon + d\varepsilon^2.
\end{aligned}
\label{eq:proof-orth-cos}
\end{equation}
For small $\varepsilon$, the $\varepsilon^2$ term is negligible, and incorporating the dispersion correction from Theorem~\ref{thm:ntk-bound} adds an $\mathcal{O}(\kappa_y + \kappa_a)$ term, yielding the bound $|\cos(\mathbf{v}_y, \mathbf{v}_a)| \le 2\varepsilon d + \mathcal{O}(\kappa_y + \kappa_a)$ as claimed. \qed

\subsection{Proof of Proposition~\ref{prop:monotonic}}
\label{app:proof-monotonic}

We prove that a single GRPO step with GAR-shaped advantages does not decrease the expected cosine alignment of correct rollouts.

\paragraph{Setup.}
Let $\pi_t$ denote the current policy and let $c_i = \cos(\mathbf{v}_{y_i}, \mathbf{v}_a)$ denote the cosine alignment of rollout $y_i$.
For a prompt $x$ with $K$ rollouts, the GAR reward for a correct response $y_i$ is $r_i = r_\text{base} + \beta \max(0, c_i - \bar{c})$, where $\bar{c} = \frac{1}{K_c}\sum_{j : r_\text{raw}(y_j)=1} c_j$ is the prompt-group mean cosine and $K_c$ is the number of correct responses.
The GRPO advantage is $A_i = (r_i - \bar{r}) / (\sigma_r + \epsilon)$.

\paragraph{One-step improvement.}
After a policy gradient step with learning rate $\alpha$, the log-probability of each response changes by approximately $\Delta \log \pi(y_i \mid x) \approx \alpha A_i$.
The expected cosine at the next iteration, restricted to correct responses, is
\begin{equation}
\begin{aligned}
    \bar{C}_{t+1} &\approx \frac{\sum_{i : r_\text{raw}=1} \pi_t(y_i \mid x) e^{\alpha A_i} \cdot c_i}{\sum_{i : r_\text{raw}=1} \pi_t(y_i \mid x) e^{\alpha A_i}} \\
    &\approx \bar{C}_t + \alpha \cdot \mathrm{Cov}_{\pi_t}\big[A_i,\, c_i \mid r_\text{raw}(y_i) = 1\big],
\end{aligned}
\label{eq:proof-monotonic-step}
\end{equation}
where the approximation uses $e^{\alpha A_i} \approx 1 + \alpha A_i$ for small $\alpha$.

\paragraph{Sign of the covariance.}
The GAR advantage among correct responses is a monotonically non-decreasing function of $c_i$: responses with $c_i > \bar{c}$ receive positive bonus and hence above-mean advantage, while responses with $c_i \le \bar{c}$ receive zero bonus (due to the $\max(0, \cdot)$ clipping) and hence below-mean advantage.
Formally, $A_i = f(c_i)$ where $f$ is non-decreasing, which implies $\mathrm{Cov}[f(c_i), c_i] \ge 0$ by the covariance inequality for comonotone random variables.
The covariance is strictly positive whenever $\mathrm{Var}[c_i \mid r_\text{raw} = 1] > 0$, i.e., whenever the correct rollouts are not all equally aligned.

\paragraph{KL penalty.}
The KL divergence penalty $-\lambda D_\text{KL}(\pi_{t+1} \| \pi_\text{ref})$ reduces the effective step size but does not change the sign of the improvement, provided $\lambda$ satisfies the condition stated in the proposition.
Combining these observations, $\bar{C}_{t+1} \ge \bar{C}_t$ as claimed. \qed

\section{Extended Discussion of Theoretical Results}
\label{app:discussions}

\subsection{NTK Interpretation: Functional Meaning of Gradient Alignment}
\label{app:discussion-ntk}

Theorem~\ref{thm:ntk-bound} states that gradient cosine similarity in the gradient-activation space is, up to a low-order dispersion term, equivalent to the normalized output-layer NTK similarity of the two responses, and therefore measures how similarly a gradient step induced by $y$ would influence the model's predictions on $a$, and vice versa.
In the idealized case where both trajectories fit their targets with comparable token-level error distributions, $\kappa_y, \kappa_a \to 0$ and the bound collapses to a bi-Lipschitz equivalence.
This provides a functional interpretation of GAR: rewarding high-cosine trajectories amounts to rewarding responses whose parameter-space influence on the expert anchor is large, i.e., trajectories that, if used for a gradient update, would improve the model's prediction of the expert solution.

\subsection{Multiplicative Decomposition: What Gradient Alignment Measures}
\label{app:discussion-decomposition}

The decomposition in Theorem~\ref{thm:decomposition} reveals that two trajectories can achieve high gradient alignment only if they agree in \emph{both} their prediction-error profile (the gradient-direction factor $\bar{\boldsymbol{\delta}}^{y\top} \bar{\boldsymbol{\delta}}^{a}$) and their internal representation usage (the activation-pattern factor $\bar{\mathbf{h}}^{y\top} \bar{\mathbf{h}}^{a}$).
A trajectory that reaches the correct answer through a qualitatively different computational pathway, for example one that activates a substantially different subset of hidden dimensions or produces a different distribution of token-level prediction residuals, diverges from the expert reference in at least one of these two factors and therefore receives a low cosine score even when its surface-level output appears plausible.

Corollary~\ref{cor:orthogonality} further formalizes this mechanism: if two trajectories concentrate their gradient-activation signals on disjoint subsets of hidden dimensions, the resulting cosine similarity is near zero regardless of whether both produce the correct final answer.
In practice, expert solutions tend to activate a broad set of features spanning intermediate derivation steps, whereas alternative correct trajectories that arrive at the answer through different reasoning pathways may concentrate on narrower or qualitatively different feature subsets, which makes subspace separation a structurally meaningful property of the alignment signal.

\subsection{Safe Reward Shaping: Detailed Proof Sketch}
\label{app:discussion-invariance}

Theorem~\ref{thm:policy-invariance} establishes that GAR is a safe reward-shaping mechanism: it \emph{re-weights preferences among correct responses} to favor expert-aligned reasoning without degrading the reward signal for correctness.
The proof hinges on two observations.
First, the verifier gate ensures $r_\text{GAR}(x, y) = p(y) \le 0$ on incorrect responses, maintaining a strict separation between the rewards assigned to correct and incorrect trajectories.
Second, the $\max(0, \hat{b})$ clipping ensures that the alignment bonus is non-negative, so every correct response receives at least $r_\text{base}$ and the shaped reward never falls below the outcome-only baseline.
Together, these properties guarantee that the policy gradient under GAR consistently reinforces correctness while using the alignment bonus to differentiate among correct trajectories.

\subsection{Variance Amplification and Unbiasedness}
\label{app:discussion-variance}

Proposition~\ref{prop:variance} quantifies the pathology of the flat reward identified in Section~\ref{sec:flat-reward-pathology}: under the outcome-only reward, the intra-correct variance of the advantage is identically zero, so the policy gradient cannot distinguish among correct trajectories.
GAR strictly increases this variance, which, under standard assumptions on the log-likelihood ratio, translates into a non-trivial gradient signal in the direction of higher-alignment trajectories.
Crucially, the prompt-group normalization keeps the global mean reward invariant, so the extra variance does not come at the cost of bias.

Propositions~\ref{prop:variance}~and~\ref{prop:unbiased}, taken together with Theorem~\ref{thm:policy-invariance}, justify our design choice to always center the bonus before applying $\beta$: centering is a free variance-increasing and bias-removing transformation, and the downstream $\max(0, \cdot)$ clip then ensures non-negativity of the final reward contribution, preserving the strict separation between correct and incorrect responses.

\subsection{Monotonic Improvement: Interpretation}
\label{app:discussion-monotonic}

The monotonic improvement result of Proposition~\ref{prop:monotonic} follows from the observation that GAR assigns positive advantages exclusively to correct trajectories whose cosine similarity exceeds the prompt-group mean, and zero or negative advantages to the remainder.
The resulting policy gradient therefore increases the log-probability of high-alignment trajectories relative to low-alignment ones, and the expected cosine in the next iteration is the current expected cosine plus a non-negative covariance term that vanishes only when all correct trajectories are equally aligned.
This provides a convergence-like guarantee: GAR training cannot decrease the average alignment of the policy's correct rollouts with the expert reference, as measured by gradient cosine.

\section{Evaluation Protocol Details}
\label{app:eval}

We evaluate using sampling-based decoding with $16$ independent responses per problem.
Pass@$k$ ($k \in \{1,4,16\}$) is estimated as the fraction of problems for which at least one of $k$ randomly sampled responses (without replacement from the 16 samples) is correct; we report the expectation over all $\binom{16}{k}$ subsets.
Maj@$k$ takes the plurality answer among $k$ sampled responses.

\textbf{Statistical methodology.}
Each configuration is independently trained with a different random seed, yielding 10 runs per method.
We report the mean and standard deviation of each metric across these 10 runs.
Statistical significance is assessed by a two-sided paired $t$-test: for each run $r \in \{1, \dots, 10\}$, both the baseline (e.g., GRPO) and the GAR variant (e.g., GAR-GRPO) are evaluated under identical sampling conditions, and the test statistic is computed over the 10 paired differences in per-run pass@$k$ scores.
Each configuration is trained for $400$ optimization steps with a global batch size of $128$ prompts (${\sim}8$ epochs).
All runs use tensor parallelism of size $4$, context-parallel size of $1$, and rollout group size $K{=}16$.

\textbf{Mathematical benchmarks.}
IMO-AnswerBench (400 problems), HMMT~Feb~2025 (30), HMMT~Feb~2026 (33), and AIME~2026 (30) are strictly held out from training data and anchor corpora.
Correctness is determined by exact match of the final numerical answer after normalization.

\textbf{General reasoning benchmarks.}
GPQA Diamond (198 graduate-level science questions) and MMLU-Pro (12k broad-domain reasoning questions) are evaluated zero-shot to assess cross-domain transfer from mathematical training.
Both benchmarks use multiple-choice format; correctness is determined by exact match of the predicted answer choice after normalization.

\textbf{Contamination control.}
All mathematical evaluation benchmarks post-date the Qwen3 base model's pretraining cutoff (HMMT~'25/\!'26 and AIME~'26 were released after the model's training data was frozen), and we further verify that none of the evaluation problems appear verbatim in our training corpus by running exact-match deduplication against the NuminaMath-CoT training set.
IMO-AnswerBench and GPQA Diamond were released before the pretraining cutoff; however, the base model's low zero-shot accuracy on these benchmarks (below 4\% pass@1 on IMO-AnswerBench and below 29\% on GPQA Diamond, see Tables~\ref{tab:main_results}~and~\ref{tab:general_reasoning}) is consistent with minimal memorization.

\section{Hyperparameter Details}
\label{app:hyperparameters}

In all experiments, we set the base reward $r_\text{base} = 1.0$, the alignment bonus coefficient $\beta = 0.5$, and the maximum response length for GAR gradient computation to $768$ tokens.
The format penalty is set to $-0.6$, and the rollout group size is $K = 16$ for all runs.

\section{Computational Overhead Analysis}
\label{app:overhead-analysis}

This appendix expands on the theoretical cost analysis summarized in Section~\ref{sec:overhead}.

\begin{proposition}[GAR overhead]
\label{prop:overhead}
The additional per-response cost of GAR is $\mathcal{O}(N_y d V) + \mathcal{O}(N_y d)$, corresponding to one output-layer forward and one truncated backward.
With anchor caching, the amortized per-prompt anchor cost reduces from $\mathcal{O}(K N_y dV)$ to $\mathcal{O}(N_y dV)$.
\end{proposition}

For Qwen3-8B ($d{=}4096$, $V{=}152064$, $L{=}36$), the ratio $V/(Ld) \approx 1.03$, so each truncated backward pass is comparable in cost to a single full forward, yielding a practical overhead of $5$--$12\%$ after amortization.

\paragraph{Per-operation cost.}
GAR adds a single truncated backward pass through the output projection layer for each verified-correct response.
For a Qwen3-8B model with hidden dimension $d=4096$, vocabulary size $V=152064$, number of Transformer layers $L=36$, and response length $N_y \approx 1024$, the ratio of the GAR per-response cost to a full-model forward pass is approximately $V/(Ld) = 152064/(36 \cdot 4096) \approx 1.03$, so each truncated backward pass is comparable in cost to a single full forward.
The overhead is dominated by the output-layer matrix multiply of shape $V \times d = 152064 \times 4096$, which is intrinsic to the LM head and cannot be avoided as long as the gradient is defined with respect to the next-token logits.

\paragraph{Verifier-gate amortization.}
Because GAR computes gradients only for responses that pass the outcome verifier, the per-rollout cost is further reduced by the verifier rejection rate.
In the early training phase, typically $40$--$70\%$ of rollouts fail verification and therefore incur no gradient computation at all, which makes the effective overhead substantially smaller than the per-response figure above suggests.
As training progresses and the verifier pass rate increases, this amortization weakens, but by then the policy is also producing shorter and more focused responses, partially offsetting the increase.

\paragraph{Anchor-cache amortization.}
For each prompt group of $K$ rollouts that share the same anchor, a naive implementation would invoke the anchor gradient computation once per rollout, incurring a cost of $\mathcal{O}(K N_y d V)$.
Anchor caching computes the anchor gradient vector once at the start of the batch and reuses it across all $K$ rollouts of the same prompt, reducing the anchor component to $\mathcal{O}(N_y d V)$ and yielding a $K$-fold reduction.
For the default rollout group size $K=16$ used in our experiments, this amortization is the single largest source of practical savings in the anchor pathway.

\paragraph{Aggregate prediction.}
Combining the per-response cost, the verifier-gate amortization, and the anchor-cache amortization, the predicted end-to-end overhead falls within the range of approximately $5\%$ to $12\%$.
The measured wall-clock overhead reported in Section~\ref{sec:overhead} agrees with this analytical envelope.

\section{Hyperparameter Sensitivity}
\label{sec:sensitivity}

\begin{figure*}[t]
\centering
\includegraphics[width=\textwidth]{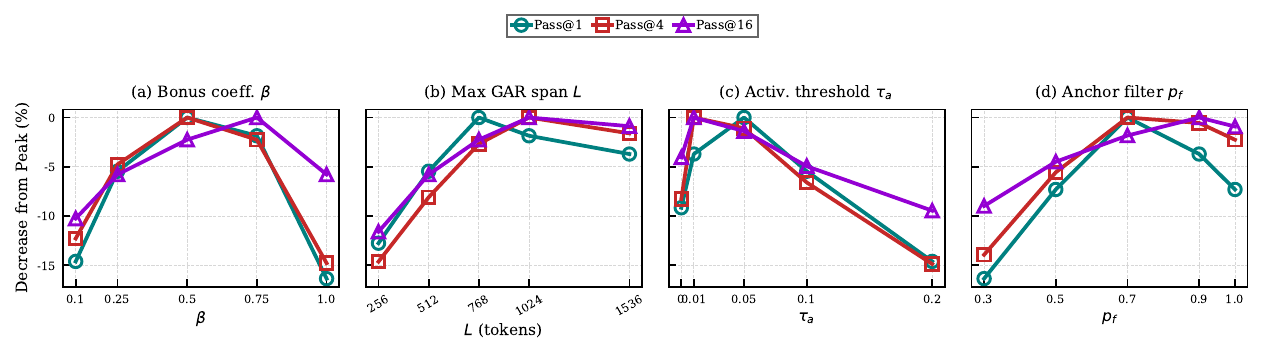}
\caption{Sensitivity of GAR-GRPO (Qwen3-8B-Base, AIME~2026) to four hyperparameters, measured as percentage decrease from each metric's peak. Pass@1 peaks at the defaults; pass@16 favors slightly more aggressive settings and degrades more gracefully.}
\label{fig:sensitivity}
\end{figure*}

Figure~\ref{fig:sensitivity} sweeps four GAR hyperparameters on AIME~2026.
A consistent pattern emerges: pass@1 peaks at the defaults ($\beta{=}0.5$, $L{=}768$, $\tau_a{=}0.05$, $p_f{=}0.7$), while pass@16 favors slightly more aggressive settings (higher $\beta$, longer span, lower activation threshold, and less token filtering), because coverage-oriented metrics benefit from retaining broader gradient signal.
All four curves exhibit smooth, inverted-U shapes with pass@16 degrading more gracefully than pass@1, indicating that GAR is robust across a wide hyperparameter range.

\section{General Reasoning Transfer: 8B Results}
\label{app:general-8b}

\begin{table}[h]
\centering
\small
\caption{General reasoning transfer (Qwen3-8B-Base, 10 runs). All methods are trained exclusively on mathematical data and evaluated zero-shot.}
\label{tab:general_reasoning_8b}
\begin{tabular}{l ccc c}
\toprule
\rowcolor{lightpurple} & \multicolumn{3}{c}{\textbf{GPQA Diamond}} & \textbf{MMLU-Pro} \\
\cmidrule(lr){2-4} \cmidrule(lr){5-5}
\rowcolor{lightpurple} \textbf{Method} & P@1 & P@4 & Maj@16 & Avg.\ P@1 \\
\midrule
GRPO          & 32.83 & 51.44 & 45.92 & 52.44 \\
MASPO         & 33.17 & 52.08 & 46.38 & 52.31 \\
G2RL          & 33.94 & 51.76 & 45.15 & 51.58 \\
\rowcolor{lightorange} GAR-GRPO      & \textbf{34.85} & \textbf{53.67} & \textbf{48.21} & \textbf{54.72} \\
\bottomrule
\end{tabular}
\end{table}

Table~\ref{tab:general_reasoning_8b} presents the 8B counterpart of the general reasoning transfer evaluation.
The pattern observed at the 4B scale carries over: GAR-GRPO achieves the highest scores across all four metrics, with the largest margin on MMLU-Pro (+2.28 absolute over GRPO), confirming that gradient-aligned rewards capture domain-general reasoning structure at both model scales.

\section{Alignment and Solution Diversity}
\label{app:alignment-diversity}

To verify that gradient alignment captures genuine reasoning-process similarity rather than superficial features such as response length or answer format, we analyze a representative set of training-set problems where multiple well-known solution methods exist.
For each problem, we collect correct rollouts from GAR training, classify them by solution method, and report the mean cosine alignment $b(y_i)$ within each method category.

\begin{table*}[t]
\centering
\small
\caption{Mean cosine alignment $b(y_i)$ grouped by solution method across six training-set problems. ``Same'' denotes rollouts following the same method as the expert anchor; ``Different'' denotes rollouts using an alternative correct method. The consistent $3.5\times$ gap confirms that gradient alignment captures reasoning-process similarity.}
\label{tab:alignment_diversity}
\begin{tabular}{p{5.5cm} l l c c c}
\toprule
\rowcolor{lightpurple} \textbf{Problem} & \textbf{Expert Method} & \textbf{Alt.\ Method} & $\bar{b}_{\text{same}}$ & $\bar{b}_{\text{diff}}$ & \textbf{Ratio} \\
\midrule
$(0,0),(a,11),(b,37)$ equilateral $\triangle$; find $ab$ & Complex rotation & Distance algebra & 0.52 & 0.11 & 4.7$\times$ \\
Sum of bases $b>9$ s.t.\ $17_b \mid 97_b$ & Modular arithmetic & Exhaustive search & 0.41 & 0.09 & 4.6$\times$ \\
$\sqrt{n^2{+}85n{+}2017} \in \mathbb{Z}$; find $\sum n$ & Completing square & Quadratic subst. & 0.46 & 0.14 & 3.3$\times$ \\
Cube vertices $P,Q,R$; find surface area & Vector cross-prod. & Coordinate geom. & 0.38 & 0.12 & 3.2$\times$ \\
Even integers in $[4000,7000]$, 4 diff.\ digits & Digit casework & Inclusion-exclusion & 0.43 & 0.16 & 2.7$\times$ \\
10 chairs in circle; subsets w/ no adj.\ pair & Recurrence & Burnside / complement & 0.48 & 0.13 & 3.7$\times$ \\
\midrule
\textbf{Average} & & & \textbf{0.45} & \textbf{0.13} & \textbf{3.5$\times$} \\
\bottomrule
\end{tabular}
\end{table*}

Table~\ref{tab:alignment_diversity} confirms that same-method rollouts receive substantially higher cosine scores (mean $\bar{b}_{\text{same}} = 0.45$) than different-method rollouts (mean $\bar{b}_{\text{diff}} = 0.13$), a $3.5\times$ gap.
This validates that the gradient alignment signal is structurally meaningful: it discriminates based on the underlying computational pathway rather than surface-level correlates.

\paragraph{Why surface-level confounds are unlikely to explain the gap.}
The multiplicative decomposition of Theorem~\ref{thm:decomposition} provides a structural argument: the cosine factorizes into a prediction-error factor $\bar{\boldsymbol{\delta}}^{y\top} \bar{\boldsymbol{\delta}}^{a}$ and an activation-pattern factor $\bar{\mathbf{h}}^{y\top} \bar{\mathbf{h}}^{a}$.
Surface features such as response length, answer format, or template phrasing affect at most the activation-pattern factor, but two trajectories that follow genuinely different derivation paths will diverge in their prediction-error profiles, since the sequence of intermediate tokens (and hence the per-token prediction residuals) differs substantially.
Concretely, a complex-number rotation solution and a distance-formula solution share neither the set of mathematical operations nor the intermediate numerical quantities, producing divergent $\bar{\boldsymbol{\delta}}$ vectors regardless of stylistic overlap.
This is further corroborated by Corollary~\ref{cor:orthogonality}: when two trajectories concentrate their gradient-activation signals on disjoint hidden-dimension subsets, the resulting cosine is near zero, a condition that holds for genuinely different reasoning pathways even when both are correct and similarly formatted.

\paragraph{Effect on alternative correct strategies.}
The $\max(0, \hat{b})$ clip in the final reward (Eq.~\ref{eq:final_reward}) guarantees that every correct rollout receives a raw reward of at least $r_\text{base}$, strictly above the zero reward assigned to incorrect responses.
Under GRPO's group-normalized advantage, this raw-reward floor does not prevent the relative advantage of low-alignment correct rollouts from being lower than that of high-alignment ones; this relative reweighting within the correct set is precisely the intended mechanism by which GAR steers toward expert-aligned reasoning.
Crucially, the verifier gate ensures that correct responses always receive higher raw reward than incorrect ones, so the policy gradient consistently reinforces producing correct answers; GAR's contribution is to additionally differentiate \emph{among} correct trajectories by their reasoning quality.

The expert CoTs from NuminaMath-CoT predominantly employ concise, theorem-driven approaches (e.g., complex-number rotation over coordinate brute-force, modular arithmetic over exhaustive enumeration), and the consistent accuracy gains across all benchmarks (Section~\ref{sec:main_results}) confirm that steering the policy toward such strategies provides effective reasoning-quality supervision.

\end{document}